\documentclass[runningheads]{llncs}
\usepackage{makeidx}
\usepackage{graphicx}
\usepackage{amsmath,amssymb} 
\usepackage{color}
\usepackage{subfigure}
\usepackage{array}
\usepackage{comment}

\usepackage{cite}

\newcolumntype{L}[1]{>{\raggedright\let\newline\\\arraybackslash\hspace{0pt}}m{#1}}
\newcolumntype{C}[1]{>{\centering\let\newline\\\arraybackslash\hspace{0pt}}m{#1}}
\newcolumntype{R}[1]{>{\raggedleft\let\newline\\\arraybackslash\hspace{0pt}}m{#1}}

\usepackage[pagebackref=true,breaklinks=true,letterpaper=true,colorlinks,bookmarks=false]{hyperref}

\begin{document}
\pagestyle{headings}
\mainmatter


\title{Discovering Latent Classes for Semi-Supervised Semantic Segmentation}

\titlerunning{Discovering Latent Classes for Semi-Supervised Semantic Segmentation}
\authorrunning{O.Zatsarynna, J.Sawatzky, J.Gall}

\makeatletter
\newcommand{\printfnsymbol}[1]{%
  \textsuperscript{\@fnsymbol{#1}}%
}
\makeatother

\newcommand\blfootnote[1]{%
  \begingroup
  \renewcommand\thefootnote{}\footnote{#1}%
  \addtocounter{footnote}{-1}%
  \endgroup
}
\author{Olga Zatsarynna\inst{1}\thanks{contributed equally} \and
Johann Sawatzky\inst{1, 2}\printfnsymbol{1} \and
Juergen Gall\inst{1}}
\institute{University of Bonn \and  EyewareTech \\
\email{\{s6olzats, jsawatzk, jgall\} @ uni-bonn.de}}

\maketitle

\begin{abstract}
High annotation costs are a major bottleneck for the training of semantic segmentation approaches. Therefore, methods working with less annotation effort are of special interest.
This paper studies the problem of semi-supervised semantic segmentation, that is only a small subset of the training images is annotated. In order to leverage the information present in the unlabeled images, we propose to learn a second task that is related to semantic segmentation but that is easier to learn and requires less annotated images. For the second task, we learn latent classes that are on one hand easy enough to be learned from the small set of labeled data and are on the other hand as consistent as possible with the semantic classes. While the latent classes are learned on the labeled data, the branch for inferring latent classes provides on the unlabeled data an additional supervision signal for the branch for semantic segmentation. 
In our experiments, we show that the latent classes boost the accuracy for semi-supervised semantic segmentation and that the proposed method achieves state-of-the-art results on the Pascal VOC 2012 and Cityscapes datasets.
\keywords{Semantic Segmentation; Semi-Supervised Learning;  Curriculum Learning; Generative Adversarial Networks.}
\end{abstract}

\section{Introduction}
\begin{figure}[t]
\centering
    \subfigure[Image]{\includegraphics[width=40mm]{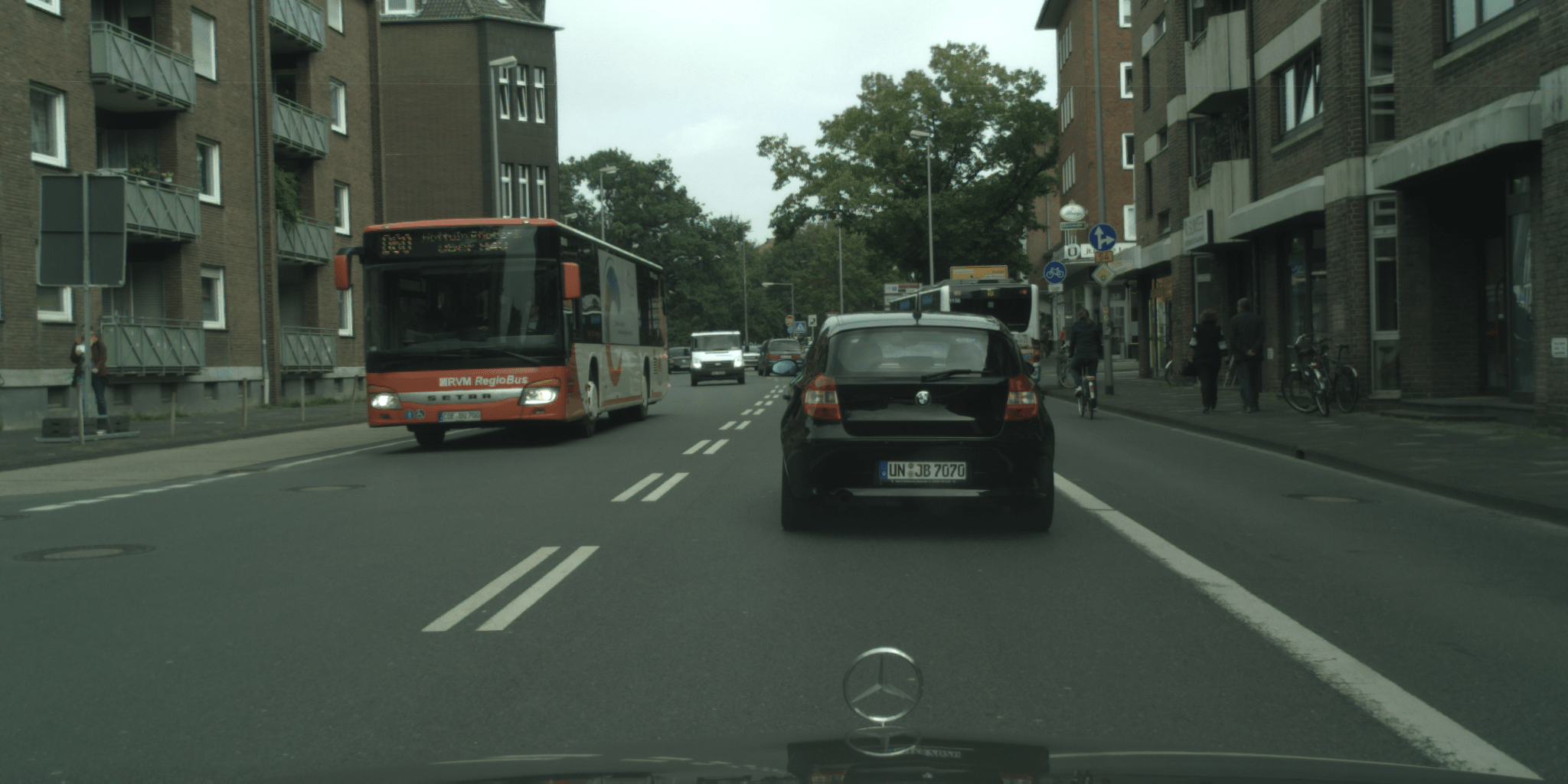}}
    \subfigure[Latent Classes]{\includegraphics[width=40mm]{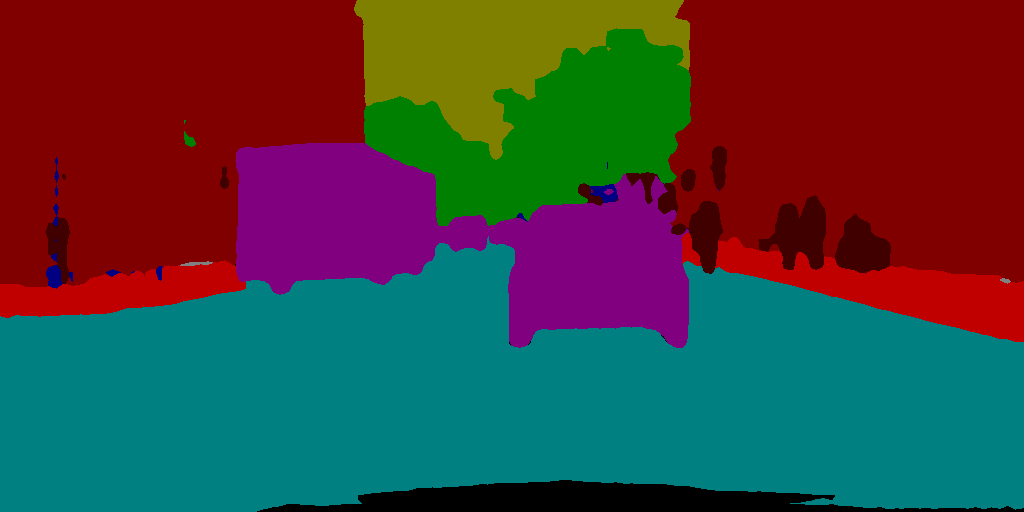}}
    \subfigure[Semantic Classes]{\includegraphics[width=40mm]{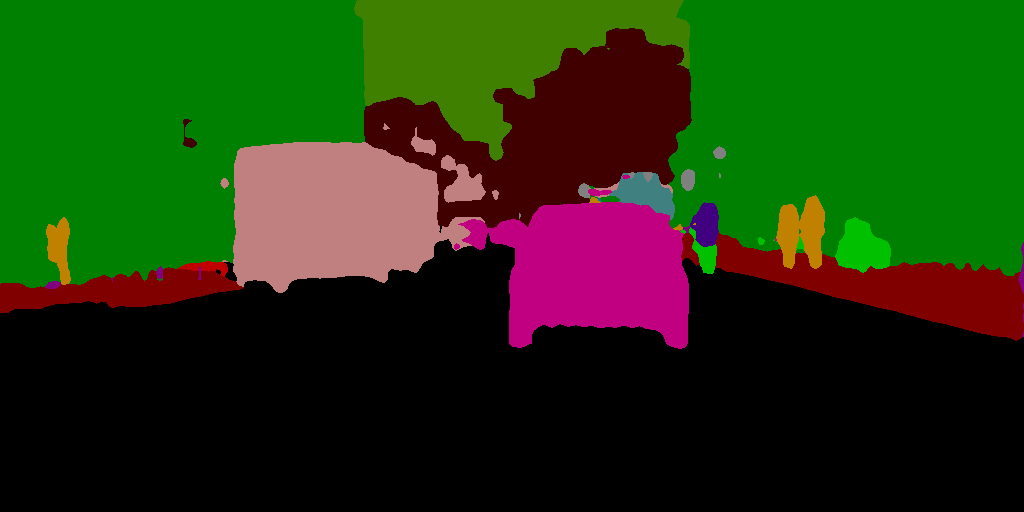}}
    \caption{Our network learns not only semantic but also latent classes that are easier to predict. The figure shows an example of latent and semantic class segmentation for an image that is not part of the training data. As it can be seen, the learned latent classes are very intuitive since the vehicles are grouped into one latent class and objects that are difficult to segment like pedestrians, bicycles, and signs are grouped into another latent class.}
\label{fig:intro}   
\end{figure}

In recent years, deep convolutional neural networks (DCNNs) have achieved astonishing performance for the task of semantic segmentation. However, to achieve good results, DCNN-based methods require an enormous amount of high-quality annotated training data and acquiring it takes a lot of effort and time. This problem is especially acute for the task of semantic segmentation, due to the need for per-pixel labels for every training image. To mitigate the annotation expenses, Hung et al.~\cite{hung2018adversarial} proposed a semi-supervised algorithm that employs images without annotation during training. On labeled data, the authors train a discriminator network that distinguishes segmentation predictions and ground-truth annotations. On unlabeled data, they use the discriminator to obtain two kinds of supervision signals. First, they use an adversarial loss to enforce realism in the predictions. Second, they use the discriminator to locate regions of sufficient realism in the prediction. These regions are then annotated by the semantic class with the highest probability. Finally, the network for semantic segmentation is trained on the labeled images and the estimated regions of the unlabeled images. Recently, Mittal et al.~\cite{mittal2019} introduced an extension to \cite{hung2018adversarial} by improving the adversarial training and adding a semi-supervised classification module. The latter is used for refining the predictions at the inference time. Although these approaches report impressive results for semi-supervised semantic segmentation, they do not leverage the entire information which is present in the unlabeled images since they discard large parts of the images. 

In this work, we propose an approach for semi-supervised semantic segmentation that does not discard any information. Our key observation is that the difficulty of the semantic segmentation task depends on the definition of the semantic classes. This means that the task can be simplified if some classes are grouped together or if the classes are defined in a different way, which is more consistent with the similarity of the instances in the feature space. If the segmentation task becomes easier, less labeled data will be required to train the network. This approach is in contrast to \cite{hung2018adversarial,mittal2019} that focus on regions in the unlabeled images that are easy to segment, whereas we focus to learn a simpler segmentation task with latent classes on the labeled data that is then used as additional guidance to learn the original task on the labeled and unlabeled data. Figure~\ref{fig:intro} shows an example of inferred latent classes and semantic classes.


Our network consists of two branches and is trained on labeled and unlabeled images jointly in an end-to-end fashion as illustrated in Figure \ref{fig:model}. While the semantic branch learns to infer the given semantic classes, the latent branch learns latent classes and infers the learned latent classes. In contrast to the semantic branch, the loss for the latent branch takes only the labeled images into account. The purpose of the latent branch is to discover latent classes that are simple enough such that they can be learned on the small set of labeled data. Without any constraints this would result in a single latent class. We therefore introduce a conditional entropy loss that minimizes the variety of semantic classes that are assigned to a particular latent class. In other words, the latent classes should be on one hand easy enough to be learned from the small set of labeled data and on the other hand they should be as consistent as possible with the semantic classes. Since the latent branch solves a simpler semantic segmentation task, we use it as additional supervision for the semantic branch on the unlabeled images. After training, the latent branch is discarded and only the semantic branch is used for inference.    

We demonstrate that our model achieves state-of-the-art results on PASCAL VOC 2012 \cite{everingham2014the} and Cityscapes \cite{cordts2016the}. 
Additionally, we show that the learned latent classes are superior to manually defined supercategories.



\section{Related Work}
The expensive acquisition of pixel-wise annotated images has been recognized as a major bottleneck for the training of deep semantic segmentation models. Consequently, the community sought ways to reduce the amount of annotated images while loosing as little performance as possible. 

Weakly-supervised semantic segmentation methods learn to segment images from cheaper image annotations, i.e.\ pixel-wise labels are exchanged for cheaper annotations for all the images in the training set. The proposed types of annotations include bounding boxes \cite{papandreou2015weakly, khoreva2017simple, li2018weakly, song2019box}, scribbles \cite{lin2016scribblesup, tang2018normalized, tang2018on} or human annotated keypoints~\cite{bearman2016what}. Image level class tags have attracted special attention. A minority of works in this area first detect potential object regions and then identify the object class using the class tags \cite{pathak2015constrained, qi2016augmented, fan2018associating}. The majority of approaches use class activation maps (CAMs) \cite{zhou2016learning} to initially locate the classes of interest. Pinheiro et al.~\cite{pinheiro2015from, shimoda2016distinct} pioneered in this area
and several methods have improved this approach \cite{kolesnikov2016seed, wei2017object, zilong2018weakly, wei2017stc, hou2018bottom, ahn2018learning, oh2017exploiting, chaudhry2017discovering, roy2017combining, wei2018revisiting, wang2018weakly, briq2018convolutional, tang2018on, ge2018multi}.
A few works leverage additional data available on the Internet. For example, \cite{hong2017weakly, jin2017webly, lee2019frame} use videos.
While the works mentioned above mainly focus on refining the localization cues obtained from the CAM, recently the task of improving the CAM itself received attention \cite{li2019tell, lee2019ficklenet, lee2019frame}.
 
Some of the works mentioned above consider a setup where some images have pixel-wise annotations and the other images are weakly labeled. They combine fully supervised learning with weakly supervised learning. Papandreou et al.~\cite{papandreou2015weakly} proposed an expectation maximization based approach, modelling the pixel-wise labels as hidden variables and the image labels or bounding boxes as the observed ones. Lee et al.~\cite{lee2019ficklenet} introduce a sophisticated dropout method to obtain better class activation maps on unlabeled images. Earlier, Li et al.~\cite{li2019tell} improved the CAMs by automatically erasing the most discriminative parts of an object. Wei et al.~\cite{wei2018revisiting} examine what improvement in CAMs can be achieved by dilated convolutions. Different from previous approaches,
Zilong et al.~\cite{zilong2018weakly} do not improve the CAM but focus on refining high confidence regions obtained from the CAM by deep seeded region growing. The semi-supervised setting without any additional weak supervision has been so far only addressed by \cite{hung2018adversarial,mittal2019}. 

While learning an easier auxiliary task as an intermediate step has been investigated in the area of domain adaptation~\cite{zhang2017curriculum, dai2019curriculum, kurmi2019curriculum, sakaridis2019guided, lian2019constructing}, it has not been studied for semi-supervised semantic segmentation. Moreover, using latent classes to facilitate learning has been investigated for object detection \cite{razavi2012latent,zhu2014capturing}, joint object detection and pose estimation \cite{li2016learning}, and weakly-supervised video segmentation \cite{richard2017weakly}. However, apart from addressing a different task, these approaches focus on discovering subcategories of classes while we aim to group the classes.

\section{Method}
\begin{figure*}[t]
\begin{center}
   \includegraphics[width=1.0\linewidth, scale=0.5]{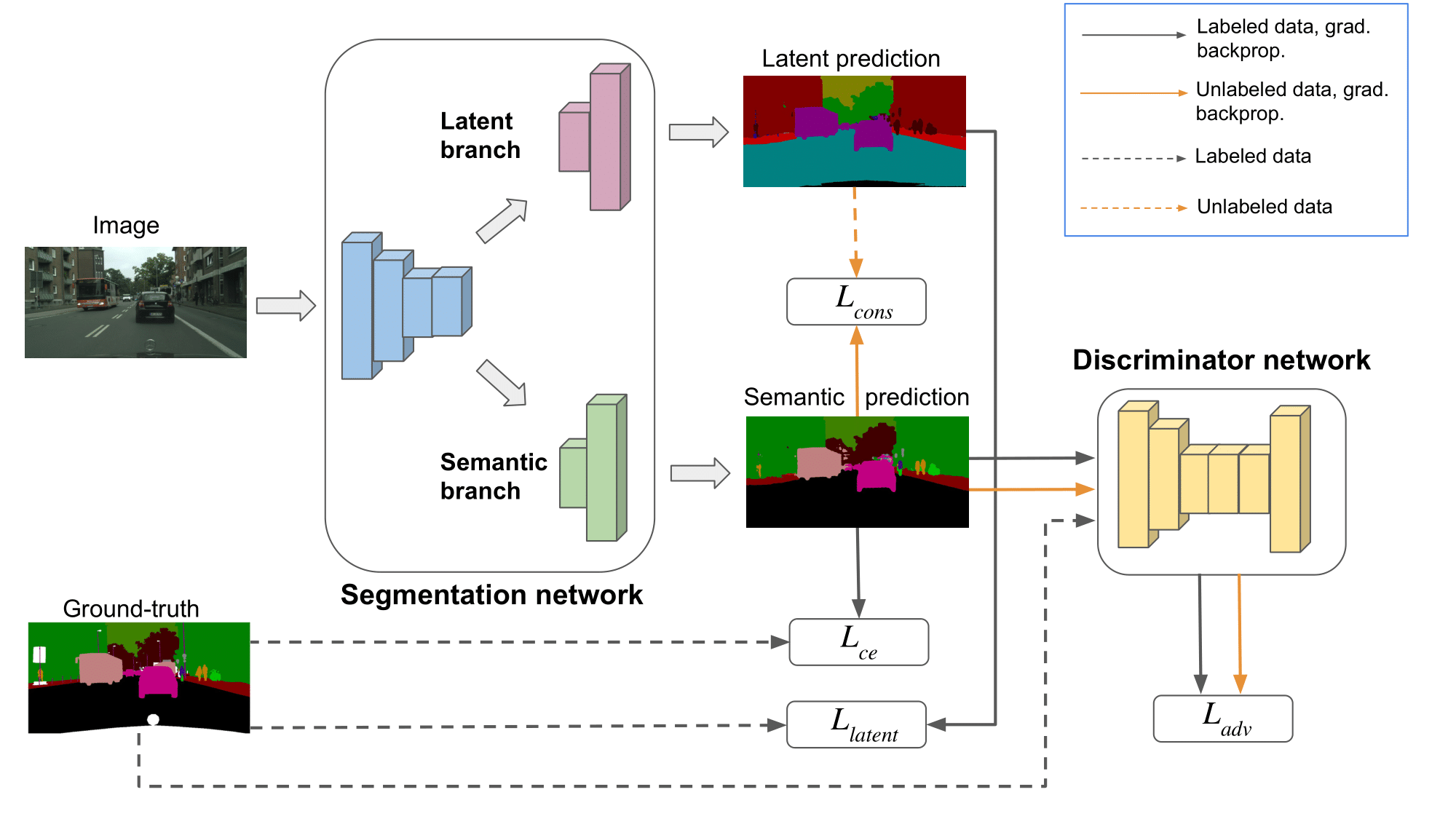}
\end{center}
   \caption{Overview of the proposed method. 
   While the semantic branch infers pixel-wise class labels, the latent branch learns latent classes and infers the learned latent classes. The latent classes are learned only on the labeled images using the latent loss $L_{latent}$ that ensures that the latent classes are as consistent as possible with the semantic classes. The semantic branch is trained on labeled images with the cross-entropy loss $L_{ce}$ and on unlabeled images the predictions of the latent branch are used as supervision ($L_{cons}$). Additionally, the semantic branch receives adversarial feedback ($L_{adv}$) from a discriminator network distinguishing predicted and ground truth segmentations.}
\label{fig:model}
\end{figure*}

An overview of our method is given in Figure~\ref{fig:model}. Our proposed model is a two-branch network. 
While the semantic branch serves to solve the final task, the purpose of the latent branch is to learn to group the semantic classes into latent classes in a data driven way as fine-grained as possible. While the fraction of annotated data is not sufficient to produce good results for the task of semantic segmentation, it is enough to learn the prediction of latent classes reasonably well, since this task is easier. Thus, the predictions of the latent branch can then serve as a supervision signal for the semantic branch on unlabeled data. 

\subsection{Semantic Branch}
\label{sec:sem}
The task of the semantic branch $S_c$ is to solve the final task of semantic segmentation, that is to predict the semantic classes for the input image. This branch is trained both on labeled and unlabeled data. 

On labeled data, we optimize the semantic branch with respect to two loss terms. The first term is the cross-entropy loss:
\begin{align}\label{eq:ce}
    L_{ce} = -\sum_{h, w, n} \sum_{c \in C} Y_n^{(h, w, c)} \log(S_{c}(X_n)^{(h, w, c)})
\end{align} where  $X_n\in\mathbb{R}^{H \times W \times 3}$ is the image, $Y_n\in\mathbb{R}^{H \times W \times |\mathcal{C}|}$ is the one-hot encoded ground truth for semantic classes, and $S_{c}$ is the predicted probability of the semantic classes.
To enforce realism in the semantic predictions, we additionally apply an adversarial loss:
\begin{equation}
\label{eq:adv}
    L_{adv} = -\sum_{n, h, w}\log(D(S_{c}(X_n))^{(h, w)})
\end{equation}
Details of the discriminator network $D$ are given in Section~\ref{sec:discriminator}

On unlabeled data, the loss function for the semantic branch also consists of two terms. The first one is the adversarial term \eqref{eq:adv} and the second term is the consistency loss that is described in Section~\ref{sec:cons}. 

\subsection{Latent Branch}
\label{sec:latent_branch}
In order to provide additional supervision for the semantic branch on the unlabeled data, we introduce a latent branch $S_l$ that is trained only on the labeled data. The purpose of the latent branch is to learn latent classes that are easier to distinguish than the semantic classes and that can be better learned on a small set of labeled images. Figure~\ref{fig:intro} shows an example of latent classes where for instance semantic similar classes like vehicles are grouped together. One of the latent classes often corresponds to a stuff class that includes all difficult classes. This is desirable since having several latent classes that are easy to recognize and one latent class that contains the rest results in a simple segmentation task that can be learned from a small set of labeled images. However, we have to prevent a trivial solution where a single latent class contains all semantic classes. We therefore propose a loss that ensures that the latent classes $l\in\mathcal{L}$ have to provide as much information about semantic classes $c\in\mathcal{C}$ as possible.                    

To this end, we use the conditional entropy as loss:
\begin{equation}\label{eq:latent}
    L_{latent} = - \sum_{l \in \mathcal{L}}\sum_{c \in \mathcal{C}} P_{b}(c,l)\log(P_{b}(c|l)).
\end{equation}
The loss is minimized if the variety of possible semantic classes for each latent class $l$ is as low as possible. In the best case, there is a one-to-one mapping between the latent and semantic classes.  
The index $b$ denotes that the probability is calculated batchwise. We first estimate the joint probability
\begin{align}
    P_{b}(c,l) = \frac{1}{NHW}\sum_{h, w, n}&S_{l}(X_n)^{(h, w, l)} Y_n^{(h, w, c)}
\end{align}
where $H$ is image height, $W$ is the image width, $N$ is the number of images in the batch, $S_{l}$ is the predicted probability of the latent classes, and $Y_n\in\mathbb{R}^{H \times W \times |\mathcal{C}|}$ is the one-hot encoded ground truth for the semantic classes.
From this, we obtain
\begin{align}
    P_{b}(c|l) = \frac{P_{b}(c,l)}{\sum_{c}P_{b}(c, l)}.
\end{align}
Obtaining the conditional entropy from multiple batches is in principle desirable, but it requires the storage of feature maps from multiple batches. Therefore we compute it per batch.

\subsection{Consistency Loss}
\label{sec:cons}
While the latent branch is trained only on the labeled data, the purpose of the latent branch is to provide additional supervision for the semantic branch on the unlabeled data. Given that the latent branch solves a simpler task than the semantic branch, we can expect that the latent classes are more accurately predicted than the semantic classes. We therefore propose a loss that measures the consistency of the prediction of the semantic branch with the prediction of the latent branch. Since the number of latent classes is less or equal than the number of semantic classes, we map the prediction of the semantic branch $S_c$ to a probability distribution of latent classes $S_{\hat{l}_c}$:
\begin{equation}
\label{eq:semantic_to_latent}
    S_{\hat{l}_c}(X_n)^{(h, w, l)} = \sum_{c \in \mathcal{C}} P(l|c) S_{c}(X_n)^{(h, w, c)}.
\end{equation}
We estimate $P(l|c)$ from the predictions of the latent branch on the labeled data. We keep track of how often semantic and latent classes co-occur with an exponentially moving average:  
\begin{equation}
    M_{c,l} ^ {(i)} = (1 - \alpha)M_{c,l} ^{(i - 1)} + \alpha \sum_{h, w, n} Y_n^{(h, w, c)} S_{l}(X_n)^{(h, w, l)}
\end{equation}
where $i$ denotes the number of the batch. The initialization is $M_{c, l}^{0} = 0$.
The parameter $0 < \alpha < 1$ controls how fast we update the average. We set $\alpha$ to the batch size divided by the number of images in the data set.
Using the acquired co-occurence matrix $M$, $P(l|c)$ is estimated as: 
\begin{align}
\label{eq:combo_matrix}
    P(l|c) = \frac{M_{c,l}}{\sum_{k \in \mathcal{L}} M_{c,k}}.
\end{align}

The consistency loss is then defined by the mean cross entropy between the latent variable maps predicted by the latent branch $S_{l}$ and the ones constructed based on the prediction of the semantic branch $S_{\hat{l}_c}$:
\begin{align}\label{eq:cons}
    L_{cons} = -\frac{1}{NHW} \sum_{n, h, w}\sum_{l \in \mathcal{L}}&S_{l}(X_n)^{(h, w, l)} \log(S_{\hat{l}_c}(X_n)^{(h, w, l)}).
\end{align}
The minimization of this loss forces the semantic branch to predict classes which are assigned to highly probable latent classes.

\subsection{Discriminator Network}
\label{sec:discriminator}

Our discriminator network $D$ is a fully-convolutional network \cite{long2015fully} with 5 layers and Leaky-ReLu as nonlinearity. It takes label probability maps from the segmentation network or ground-truth maps as input and predicts spatial confidence maps. Each pixel represents the confidence of the discriminator about whether the corresponding pixel in a semantic label map was sampled from the ground-truth map or the segmentation prediction. 
We train the discriminator network with help of the spatial cross-entropy loss using both labeled and unlabeled data:
\begin{align}
    L_D &= -\sum_{h, w} (1 - y_n) \log(1 - D(S_{c}(X_n))^{h, w}) + y_n\log(D(Y_n)^{h, w})
\end{align}
where $y_n = 0$ if a sample is drawn from the segmentation network, and $y_n = 1$ if it is a ground-truth map. 
By minimizing such a loss, the discriminator learns to distinguish between the generated and ground-truth probability maps.

\section{Experiments}

\subsection{Implementation Details}
\label{sec:implementation}
For a fair comparison with Hung et al.~\cite{hung2018adversarial} and Mittal et al.~\cite{mittal2019}, we choose the same backbone architecture and keep the same hyper-parameters where appropriate. 
For the segmentation network, we use a single scale ResNet-based DeepLab-v2 \cite{chen2018deeplab} architecture that is pre-trained on the ImageNet \cite{russakovsky2015imagenet} and MSCOCO \cite{lin2014microsoft}. We branch the proposed network at the last layer by applying Atrous Spatial Pyramid Pooling (ASPP) \cite{chen2018deeplab} two times for the semantic and latent branch. Finally, we use bilinear upsampling to make the predictions match the initial image size.  

For the discriminator network, we use a fully convolutional network, which contains 5 convolutional layers with kernels of the sizes $4 \times 4$ and 64, 128, 256, 512 and 1 channels, applied with a stride equal to 2. Each convolutional layer, except for the last one, is followed by a Leaky-ReLU with the leakage coefficient equal to 0.2.

We train the segmentation network on labeled and unlabeled data jointly with $L = L_{labeled} + 0.1 \cdot L_{unlabeled}$ where the weight factor is the same as in \cite{hung2018adversarial}. The loss for the labeled and unlabeled data are given by 
\begin{align}
    &L_{labeled} = L_{ce} + L_{latent} + 0.01 \cdot L_{adv},\\
    &L_{unlabeled} = L_{cons} + 0.01 \cdot L_{adv}.
\end{align}
The weight for the adversarial loss is also the same as in \cite{hung2018adversarial}. By default, we limit the number of latent classes to 20. Additional details are provided as part of the supplementary material.




We conducted our experiments on three datasets for semantic segmentation: Pascal VOC 2012 \cite{everingham2014the}, Cityscapes \cite{cordts2016the} and IIT Affordances \cite{nguyen2018affordances}. We report the results for the IIT Affordances dataset \cite{nguyen2018affordances} in the supplementary material.
The Pascal VOC 2012 dataset contains images with objects from 20 foreground classes and one background class. There are 10528 training and 1449 validation images in total. The testing of the resulting model is carried out on the validation set.
The Cityscapes dataset comprises images extracted from 50 driving videos. It contains 2975, 500 and 1525 images in the training, validation and test set, respectively, with annotated objects from 19 categories. We report the results of testing the resulting model on the validation set.
As an evaluation metric, we use mean-intersection-over-union (mIoU).

\subsection{Comparison with the State-of-the-Art}
\begin{table}[t]
      \caption{Comparison to the state-of-the-art on Pascal VOC 2012 using mIoU (\%). }
      \centering
    \begin{tabular}{|C{35mm}|C{10mm}|C{10mm}|C{10mm}|C{10mm}|C{10mm}|C{10.5mm}|}
    \hline
    & \multicolumn{6}{c|}{Fraction of annotated images} \\ \hline
    Method & 1/50 & 1/20 & 1/8 & 1/4 & 1/2 & Full  \\ \hline
    \hline
    Hung et al.~\cite{hung2018adversarial} & 55.6 & 64.6 & 69.5 & 72.1 & 73.8  & 74.9 \\ \hline
    Mittal et al.~\cite{mittal2019} & \textbf{63.3} & 67.2 & 71.4 & - & - & {75.6} \\ \hline
    Proposed & 59.6 & 68.2 & 71.3  & \textbf{72.4}  & \textbf{73.9}  & 75.0 \\ \hline
    Proposed + Classifier & 61.8 & \textbf{69.3} & \textbf{72.2} & - & - & 75.3 \\ \hline
    \end{tabular}
    \label{tab:result_VOC}
\end{table}
\subsubsection{PASCAL VOC 2012.}
On the PASCAL VOC 2012 dataset, we conducted our experiments on five fractions of annotated images, as shown in Table~\ref{tab:result_VOC}, where the rest of the images are used as unlabeled data. Since \cite{hung2018adversarial} report the results only for the latest three fractions, we evaluate the performance of their method for the unreported fractions based on the publicly available code.
The improvement is especially pronounced, if we look at the sparsely labeled data fractions, such as $1/50$, $1/20$ and $1/8$. 
Our method performs on par with \cite{mittal2019} and the leading method varies from data fraction to data fraction. However, our approach of learning latent variables is complementary to \cite{mittal2019} and we can also add a classifier for refinement as in \cite{mittal2019}.   
We show some qualitative results of our method in the supplementary material.

\begin{table}[t]
    \caption{Comparison to the state-of-the-art on Cityscapes using mIoU (\%).}
    \centering
    \begin{tabular}{|C{30mm}|C{20mm}|C{12mm}|C{12mm}|C{12mm}|C{12.5mm}|}
    \hline
    \multicolumn{2}{|c}{} & \multicolumn{4}{|c|}{Fraction of annotated images} \\ \hline
    Method & Pre-training & 1/8 & 1/4 & 1/2 & Full  \\ \hline
    \hline
    Mittal et al.~\cite{mittal2019} & & 59.3 & 61.9 & - & 65.8 \\ \hline 
    Proposed & & 61.0 & 63.1 & - & 64.9 \\ \hline
    Hung et al.~\cite{hung2018adversarial} & COCO  & 58.8  & 62.3 & 65.7 & 67.7 \\ \hline
    Proposed & COCO &\textbf{63.3}  & \textbf{65.4} & \textbf{66.1}  & 66.3 \\ \hline
    \end{tabular}
    \smallskip
    \label{tab:result_CS}
\end{table}


\subsubsection{Cityscapes.}
For the Cityscapes dataset, we follow the semi-supervised learning protocol that was proposed in \cite{hung2018adversarial}. This means that $1/8$, $1/4$ or $1/2$ of the training images are annotated and the other images are used without any annotations.
We report the results in Table~\ref{tab:result_CS}. Since \cite{mittal2019} does not pre-train the segmentation network on COCO, we evaluated our method also without COCO pre-training. We outperform both \cite{hung2018adversarial} and \cite{mittal2019} on all annotated data fractions.  
We show some qualitative results of our method in the supplementary material.

\subsection{Ablation Experiments}
In our ablation experiments, we evaluate the impact of each loss term. Then we examine the impact of the number of latent classes and show that they form meaningful supercategories of the semantic classes. Finally, we show that the learned latent classes outperform supercategories that are defined by humans.

\begin{table}[t]
\centering
\caption{Impact of the loss terms. The evaluation is performed on Pascal VOC 2012 where 1/8 of the data is labeled. $L^{labeled}_{adv}$ denotes that the adversarial loss is only used for the labeled images.}
\begin{tabular}{|L{50mm}|C{20mm}|}
\hline
Loss & mIoU (\%)  \\ \hline \hline
$L_{ce}$ & 64.1 \\ \hline
$L_{ce} + L_{latent}$ & 64.6 \\ \hline
$L_{ce} + L_{latent} + L_{cons}$ & 67.3 \\ \hline
$L_{ce} + L^{labeled}_{adv}$ & 68.7 \\ \hline
$L_{ce} + L_{adv}$ & 69.4 \\ \hline 
$L_{ce} + L_{latent} + L_{cons} + L_{adv}$ & 71.3 \\ \hline
\end{tabular} \smallskip
\label{tab:components}
\end{table}

\subsubsection{Impact of the loss terms.} For analyzing the impact of the loss terms $L_{ce}$~\eqref{eq:ce}, $L_{adv}$~\eqref{eq:adv}, $L_{latent}$~\eqref{eq:latent}, and $L_{cons}$~\eqref{eq:cons}, we use the Pascal VOC 2012 dataset where 1/8 of the data is labeled. The results for different combinations of loss terms are reported in Table~\ref{tab:components}.

We start using only the entropy loss $L_{ce}$ since this loss is always required. In this setting only the semantic branch is used and trained only on the labeled data. This setting achieves 64.1\% mIoU. Adding the latent loss $L_{latent}$ improves the performance by 0.5\%. In this setting, the semantic and latent branch are used, but they are both only trained on the labeled data. Adding the consistency loss $L_{cons}$ boosts the accuracy by 2.7\%. This shows that the latent branch provides additional supervision for the semantic branch on the unlabeled data. 

So far, we did not use the adversarial loss $L_{adv}$. When we add the adversarial loss only for the labeled data $L^{labeled}_{adv}$ to the entropy loss $L_{ce}$, the performance grows by 4.6\%. In this setting, only the labeled data is used for training. If we use the adversarial loss also for the unlabeled data, the accuracy increases by 0.7\%. This shows that adversarial loss improves semi-supervised learning, but the gain is not as high compared to additionally using the latent branch to supervise the semantic branch on the unlabeled data. In this setting, all loss terms are used and the accuracy increases further by 1.9\%. Compared to the entropy loss $L_{ce}$, the proposed loss terms increase the accuracy by 7.2\%.    

\subsubsection{Impact of number of latent classes.}
For our approach, we need to specify the maximum number of latent classes. While we used by default 20 in our previous experiments, we now evaluate it for 2, 4, 6, 10, and 20 latent classes on Pascal VOC 2012 with 1/8 of the data being labeled. The results are reported in Table~\ref{tab:numlv}. The performance grows monotonically with the number of latent classes reaching its peak for 20. 

In the same table, we also report the number of effective latent classes. We consider a latent class $l$ to be effectively used at threshold $t$, if $P(l|c) > t$ for at least one semantic class $c$. We report this number for $t=0.1$ and $t=0.9$. The number of effective latent classes differs only slightly for these two thresholds.  This shows that a latent class typically either constitutes a supercategory of at least one semantic class or it is not used at all. We observe that until 10, all latent classes are used. If we allow up to 20 latent classes, only 14 latent classes are effectively used. In practice, we recommend to set the number of maximum latent classes to the number of semantic classes. The approach will then select as many latent classes as needed. Although we assume that the number of latent classes is less or equal to the number of semantic classes, we also evaluated the approach for 50 latent classes. As expected, the accuracy drops but the approach remains stable. The number of effectively used latent classes also remains at 14. In practice, this setting should not be used since it violates the assumptions of the approach and can lead to unexpected behavior in some cases.


\begin{table}[t]
\centering
\caption{Impact of number of latent classes. The evaluation is performed on Pascal VOC 2012 where 1/8 of the data is labeled. A latent class $l$ is considered effective, if there exists a semantic class $c$ so that $P(l|c) > t$. The third column shows this number for $t=0.1$ and the fourth for $t=0.9$.}
\begin{tabular}{|C{30mm}|C{20mm}|C{20mm}|C{20mm}|}
\hline
\multicolumn{2}{|c|}{} & \multicolumn{2}{|c|}{Effective latent classes}  \\ \hline
Max. latent classes & mIoU (\%) & $t=0.1$ & $t=0.9$  \\ \hline
\hline
2 & 69.7 & 2& 2\\ \hline
4 & 70.2 & 4& 4\\ \hline
6 & 70.3 & 6& 6\\ \hline
10 & 70.7 & 10& 10\\ \hline
20 & 71.3 & 16& 14\\ \hline
50 & 70.8 & 18& 14\\ \hline
\end{tabular} \smallskip
\label{tab:numlv}
\end{table}


\begin{figure}[t!]
\centering
\subfigure[$P(l|c)$ on Pascal VOC 2012 for 10 and 20 latent classes]{\includegraphics[width=.45\linewidth, height=35mm]{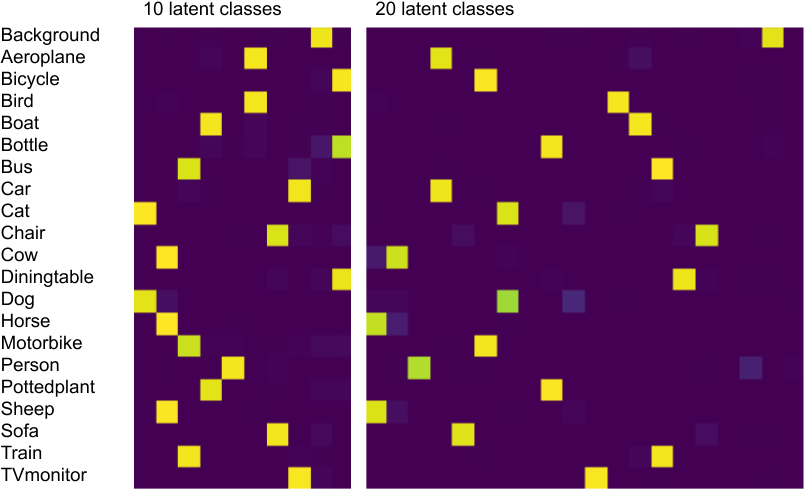}}
\subfigure[$P(l|c)$ on Cityscapes for 10 and 20 latent classes. ]{\includegraphics[width=.45\linewidth, height=35mm]{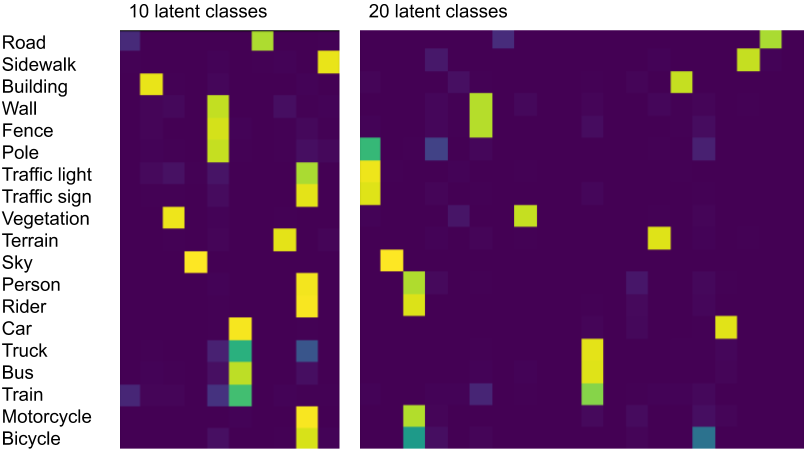}}
 \caption{The distribution of latent classes for both datasets is pretty sparse, essentially the latent classes form supercategories of semantic classes that are similar in appearance. 
 The grouping bicycle, bottle, and dining table for 10 latent classes seems to be unexpected, but due to the low number of latent classes the network is forced to group additional semantic classes. In this case, the network tends to group the most difficult classes of the dataset. However, we observed that there are small variations of the groupings for different runs when the number of latent classes is too small. In case of 20 latent classes, the merged classes are very intuitive, but not all latent classes are effectively used.}
  \label{fig:combos}
\end{figure}


To see if a semantic class is typically mapped to a single latent class, we plot $P(l|c)$ for inference on Pascal VOC 2012 as well as on Cityscapes and show the results in Figure~\ref{fig:combos}(a) and Figure~\ref{fig:combos}(b), respectively. Indeed, the mapping from semantic classes to latent classes is very sparse. Typically, for each semantic class $c$, there is one dominant latent class $l$, i.e., $P(l|c)>0.9$. If the number of latent classes increases to 20, some of the latent classes are not used. On Pascal VOC 2012, similar categories like cat and dog or cow, horse, and sheep are grouped. Some groupings are based on the common background like aeroplane and bird. The grouping bicycle, bottle, and dining table combines the most difficult classes of the dataset. On Cityscapes with 20 latent classes, the semantic classes pole, traffic light, and traffic sign; person, rider, motorcycle, and bicycle; wall and fence; truck, bus, and train are grouped together. These groupings are very intuitive. 

\subsubsection{Comparison of learned latent classes with manually defined latent classes.}
Since the latent classes typically learn supercategories of the semantic classes, the question arises if the same effect can be achieved with manually defined supercategories. In this experiment, the latent classes are replaced with 10 manually defined supercategories. More details regarding these  supercategories are provided in the supplementary material. In this setting, the latent branch is trained to predict these supercategories on the labeled data using the cross-entropy loss. For unlabeled data, everything remains the same as for the proposed method. We report the results in Table~\ref{tab:hcc}. The performance using the supercategories is only 69.0\%, which is significantly below the proposed method for 10 latent variables.

Another approach would be to learn all semantic classes instead of the latent classes in the latent branch. In this case, both branches learn the same semantic classes. This gives 68.5\%, which is also worse than the learned latent classes. If both branches predict the same semantic classes, we can also train them symmetrically. Being more specific, on labeled data they are both trained with the cross-entropy loss as well as the adversarial loss. On unlabeled data, we apply the adversarial loss to both of them and use the symmetric Kullback–Leibler divergence (KL) as a consistency loss. This approach performs better, giving 69.1\%, but it is still inferior to our proposed method. Overall, this shows the necessity to learn the latent classes in a data-driven way.

\begin{table}[t]
\centering
\caption{Comparison of learned latent classes with manually defined latent classes. The evaluation is performed on Pascal VOC 2012 where 1/8 of the data is labeled. In case of learned latent classes, the second column reports the maximum number of latent classes. In case of manually defined latent classes, the exact number of classes is reported. } 
\begin{tabular}{|L{33mm}|C{20mm}|C{20mm}|}
\hline
Method & Classes & mIoU (\%)  \\ \hline \hline
Manual & 10 & 69.0 \\ \hline
Learned & 10 & {70.7} \\ \hline
\hline
Semantic classes & 21 & 68.5  \\ \hline
Semantic classes (KL) & 21& 69.1  \\ \hline
Learned & 20 & {71.3} \\ \hline
\end{tabular} \smallskip
\label{tab:hcc}
\end{table}

\section{Conclusion}
In this work, we addressed the task of semi-supervised semantic segmentation, where a small fraction of the data set is labeled in a pixel-wise manner, while most images do not have any types of labeling. Our key contribution is a two-branch segmentation architecture, which uses latent classes learned in a data-driven way on labeled data to supervise the semantic segmentation branch on unlabeled data. We evaluated our approach on the Pascal VOC 2012 and the Cityscapes dataset where the proposed method achieves state-of-the-art results.

\section*{Acknowledgement}
This work was funded by the Deutsche Forschungsgemeinschaft (DFG, German Research Foundation) under Germany’s Excellence Strategy – EXC 2070 – 390732324.

\bibliographystyle{splncs03}
\bibliography{egbib}

\title{Supplementary Material: Discovering Latent Classes for Semi-Supervised Semantic Segmentation}
\titlerunning{Discovering Latent Classes for Semi-Supervised Semantic Segmentation}
\authorrunning{O.Zatsarynna, J.Sawatzky, J.Gall}
\author{Olga Zatsarynna\inst{1}\thanks{contributed equally} \and
	Johann Sawatzky\inst{1, 2}\printfnsymbol{1} \and
	Juergen Gall\inst{1}}
\institute{University of Bonn \and  EyewareTech \\
	\email{\{s6olzats, jsawatzk, jgall\} @ uni-bonn.de}}	
\maketitle
	
\section{Training Details}
	The optimization of the segmentation network is performed using SGD with a momentum equal to 0.9 and the learning rate decay of $10^{-4}$. The learning rate, that is initially equal to $2.5 \cdot 10^{-4}$, is decreased with polynomial decay with the power of $0.9$.
	For the discriminator, we employ the Adam optimizer \cite{kingma2014adam}, where the initial learning rate is equal to $10^{-4}$ and that follows the same decay schedule as introduced for the segmentation network.
	
	At each iteration, we alternately apply the described training scheme on the batch of the randomly sampled labeled and unlabeled data. To ensure the robustness of the evaluation procedure, we report results averaged over 5 random seeds that control the sampling procedure.
	We add the consistency loss term only after 5000 iterations since the latent branch needs to learn some useful latent classes first.
	
	On Pascal VOC 2012, during the training procedure, the images are cropped with crop size equal to $321 \times 321$ and undergo random scaling and horizontal mirroring. 
	We train our model for 20k iterations with a batch size of 10 images. The testing of the resulting model is carried out on the validation set.
	
	On the Cityscapes dataset, during training, we pre-process the images by performing cropping operations with crop size equal to $505 \times 505$ and additionally apply random scaling and horizontal mirroring.
	On the Cityscapes dataset, our model is trained for 40k iterations with batches of size 2. We report the results of testing the resulting model on the validation set.
	
	\section{IIT Affordances}
	The IIT Affordances dataset \cite{nguyen2018affordances} contains images of 10 common human tools.  It has 8835 images in total, where 50\% are used for the training split, 20\% for the validation split, and the rest 30\% for the test split. Around 60\% of the images in the dataset are from ImageNet, while the rest are taken from cluttered scenes, which implies a large variation of images within the dataset.
	
	During training, the images are cropped with the crop size equal to $321 \times 321$ and undergo random scaling and horizontal mirroring. 
	We train our model for 20k iterations with a batch size of 10 images on the training and validation images together. The testing of the resulting model is carried out on the test set. We report the results in the Table~\ref{tab:result_IIt}. As for the other datasets, our approach outperforms \cite{hung2018adversarial}.
	
	\begin{table}[t]
		\centering
		\caption{Comparison to Hung et.al on IIT Affordances. We used 7 latent classes for the proposed model}
		\begin{tabular}{|C{25mm}|C{12mm}|C{12mm}|C{12mm}|}
			\hline
			\multicolumn{4}{|c|}{IIT 2017 Affordances}  \\ \hline
			& \multicolumn{3}{c|}{Fraction of annotated images} \\ \hline
			Method & 1/50 & 1/20 & 1/8  \\ \hline
			& \multicolumn{3}{c|}{mIoU (\%)} \\ \hline
			\hline
			Hung et al.\cite{hung2018adversarial} & 47.4  & 55.8 & 64.3 \\ \hline
			Proposed &\textbf{51.3}  & \textbf{58.8} & \textbf{65.4} \\ \hline
		\end{tabular}
		\smallskip
		\label{tab:result_IIt} 
	\end{table}

	\section{Qualitative Results}
\begin{figure*}
		\centering
		\subfigure{\includegraphics[width=18mm]{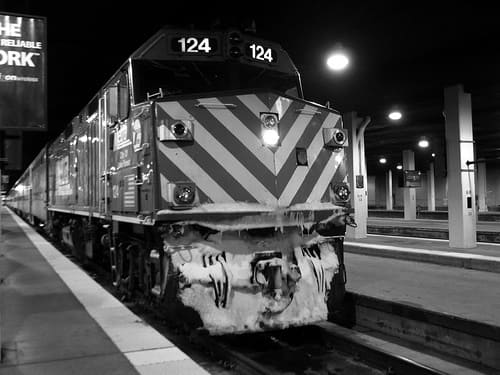}}~~
		\subfigure{\includegraphics[width=18mm]{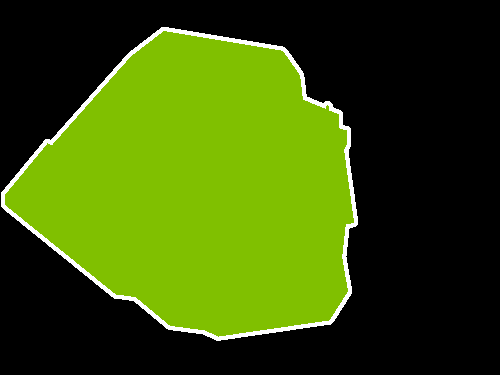}}~~
		\subfigure{\includegraphics[width=18mm]{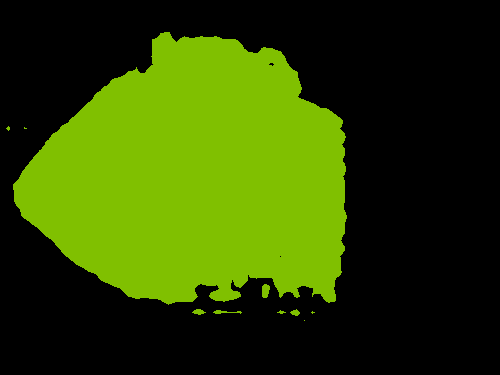}}~~
		\subfigure{\includegraphics[width=18mm]{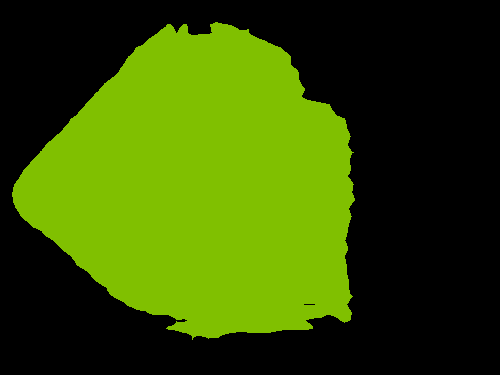}}~~
		\subfigure{\includegraphics[width=18mm]{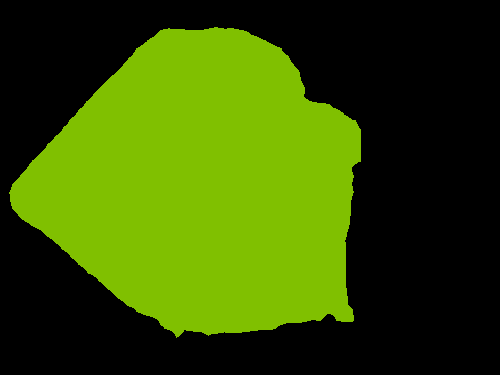}}~~
		\subfigure{\includegraphics[width=18mm]{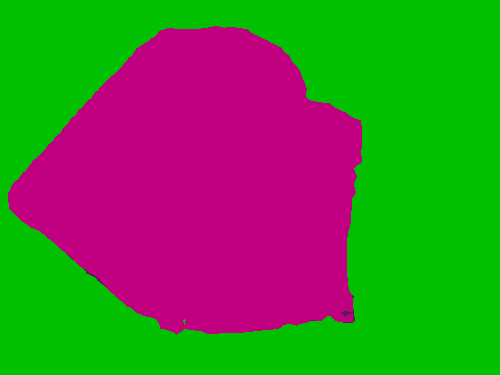}}~~\\
		\subfigure{\includegraphics[width=18mm]{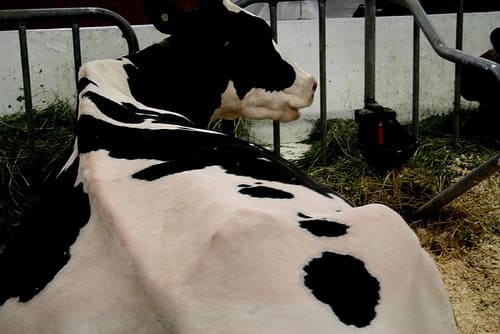}}~~
		\subfigure{\includegraphics[width=18mm]{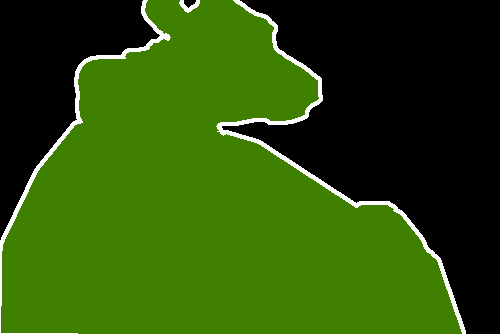}}~~
		\subfigure{\includegraphics[width=18mm]{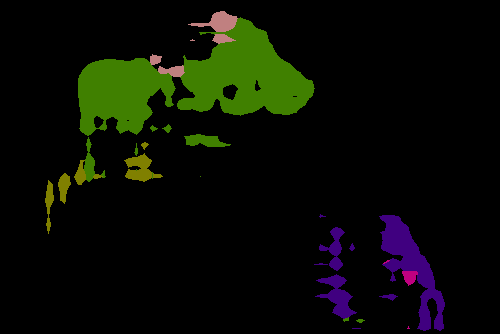}}~~
		\subfigure{\includegraphics[width=18mm]{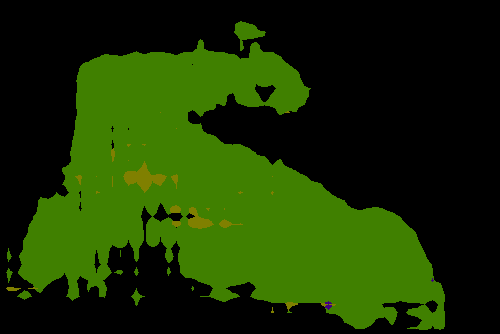}}~~
		\subfigure{\includegraphics[width=18mm]{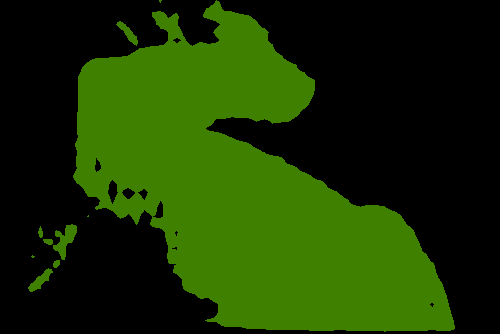}}~~
		\subfigure{\includegraphics[width=18mm]{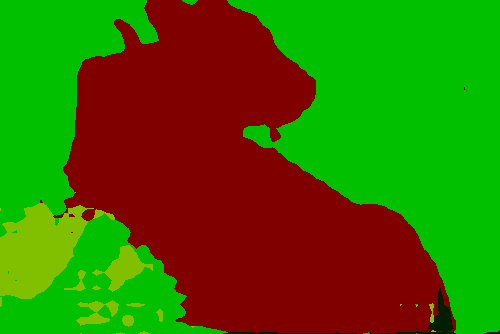}}~~\\
		\subfigure{\includegraphics[width=18mm]{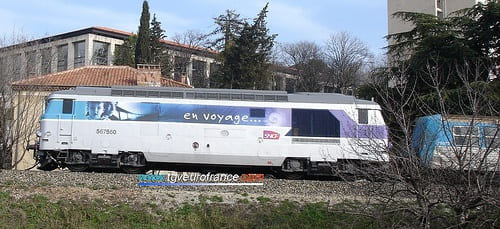}}~~
		\subfigure{\includegraphics[width=18mm]{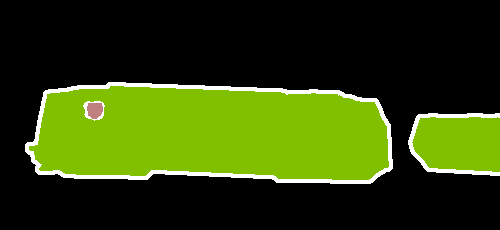}}~~
		\subfigure{\includegraphics[width=18mm]{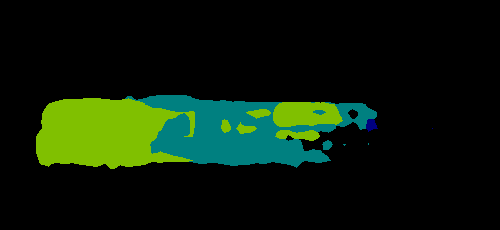}}~~
		\subfigure{\includegraphics[width=18mm]{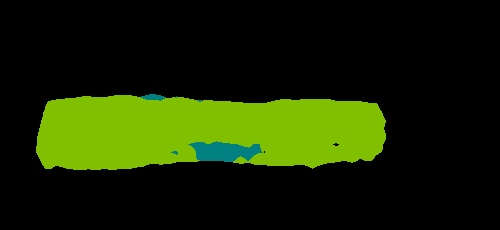}}~~
		\subfigure{\includegraphics[width=18mm]{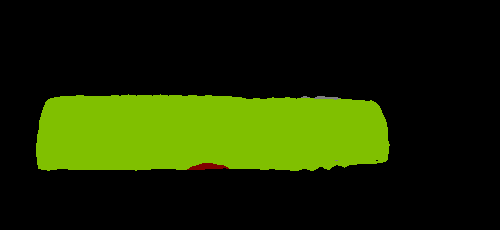}}~~
		\subfigure{\includegraphics[width=18mm]{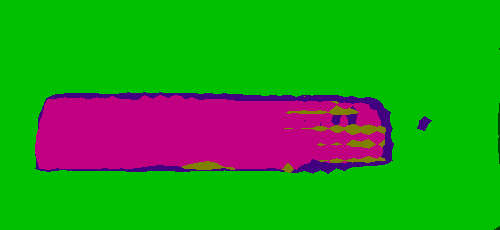}}~~\\
		\subfigure{\includegraphics[width=18mm]{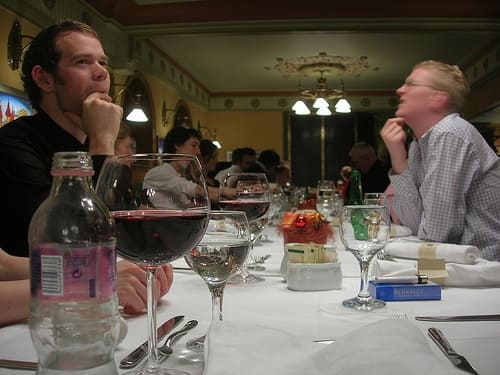}}~~
		\subfigure{\includegraphics[width=18mm]{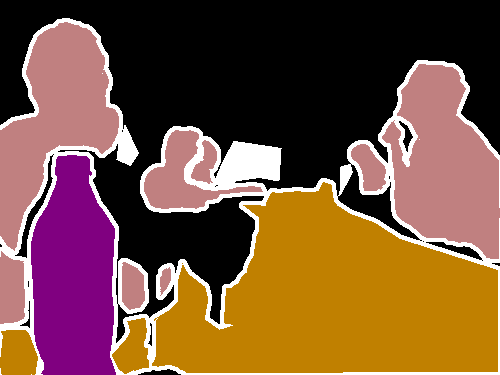}}~~
		\subfigure{\includegraphics[width=18mm]{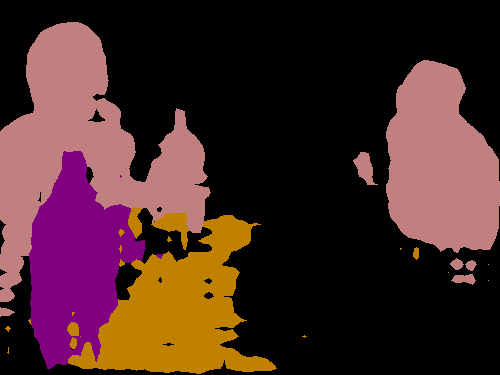}}~~
		\subfigure{\includegraphics[width=18mm]{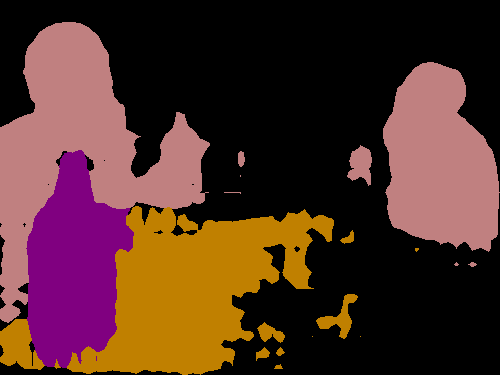}}~~
		\subfigure{\includegraphics[width=18mm]{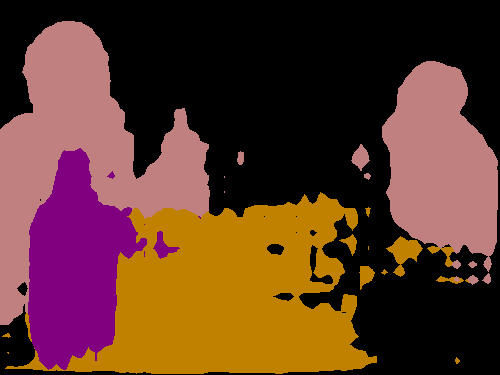}}~~
		\subfigure{\includegraphics[width=18mm]{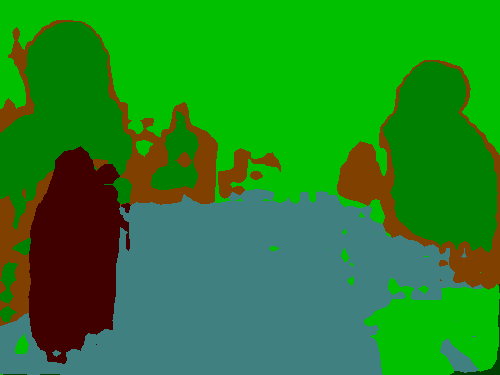}}~~\\
		\subfigure{\includegraphics[width=18mm]{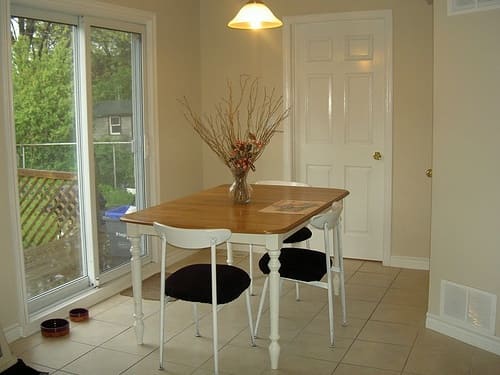}}~~
		\subfigure{\includegraphics[width=18mm]{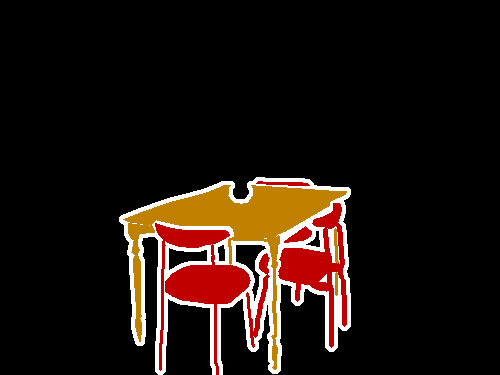}}~~
		\subfigure{\includegraphics[width=18mm]{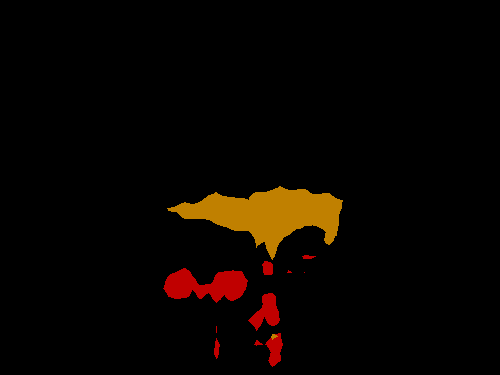}}~~
		\subfigure{\includegraphics[width=18mm]{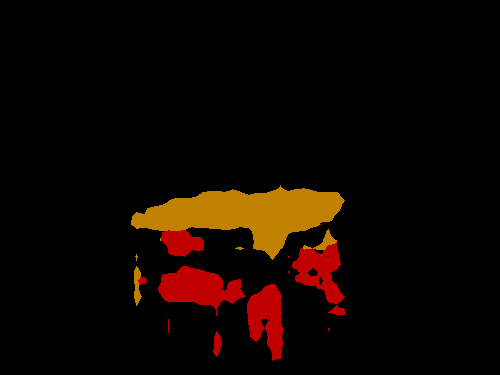}}~~
		\subfigure{\includegraphics[width=18mm]{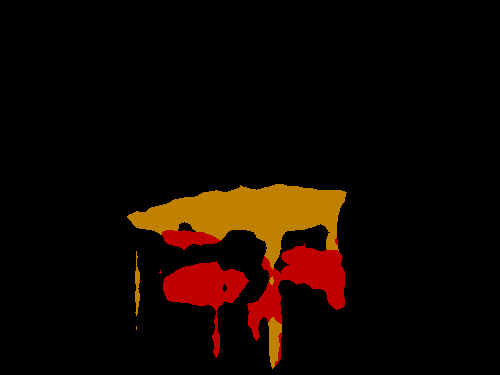}}~~
		\subfigure{\includegraphics[width=18mm]{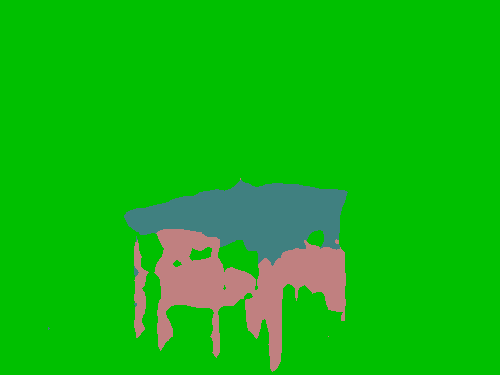}}~~\\
		\subfigure{\includegraphics[width=18mm]{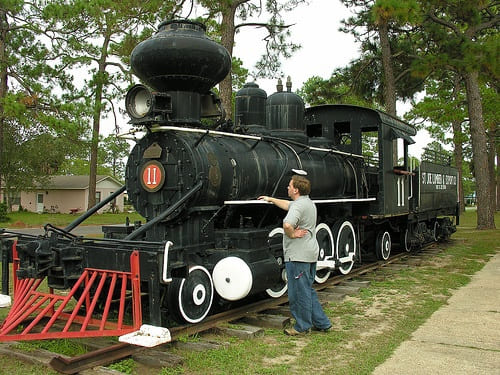}}~~
		\subfigure{\includegraphics[width=18mm]{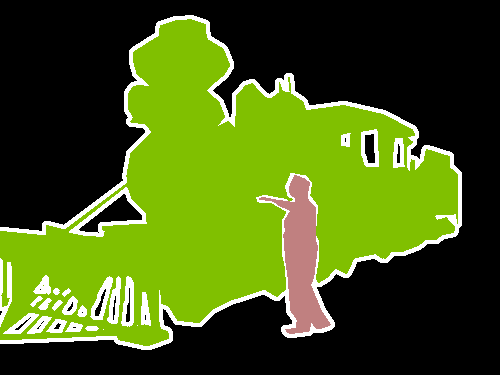}}~~
		\subfigure{\includegraphics[width=18mm]{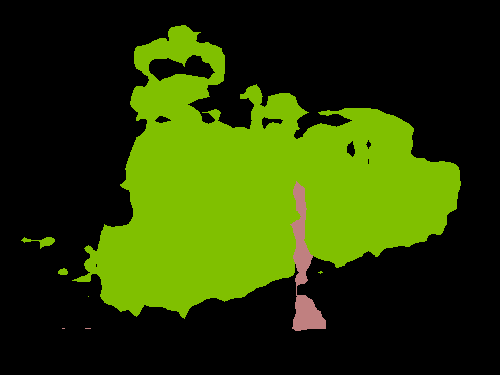}}~~
		\subfigure{\includegraphics[width=18mm]{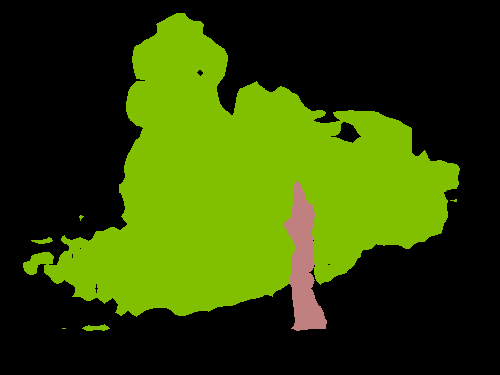}}~~
		\subfigure{\includegraphics[width=18mm]{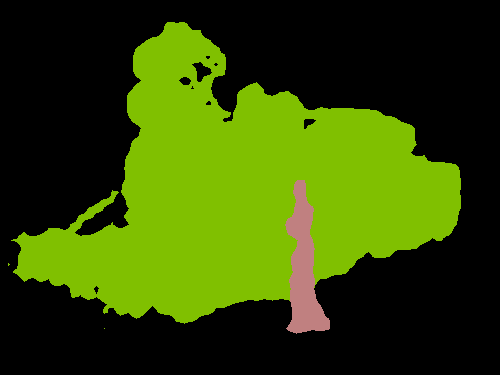}}~~
		\subfigure{\includegraphics[width=18mm]{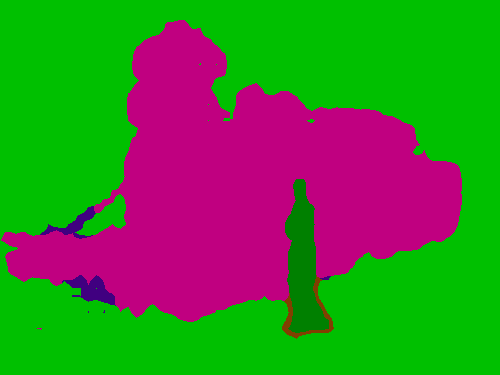}}~~\\
		\subfigure{\includegraphics[width=18mm]{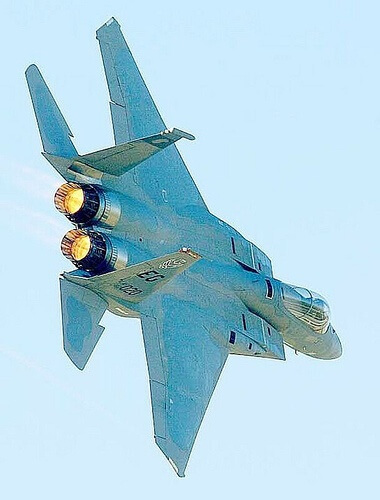}}~~
		\subfigure{\includegraphics[width=18mm]{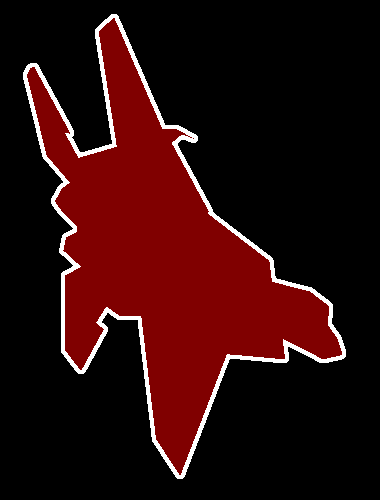}}~~
		\subfigure{\includegraphics[width=18mm]{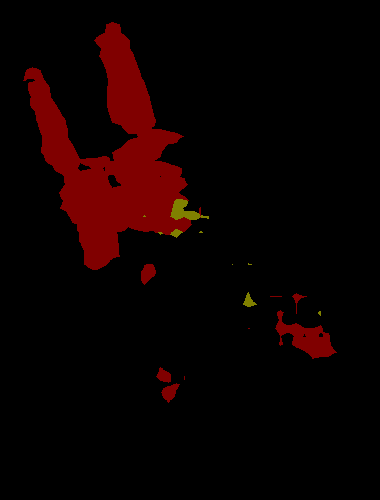}}~~
		\subfigure{\includegraphics[width=18mm]{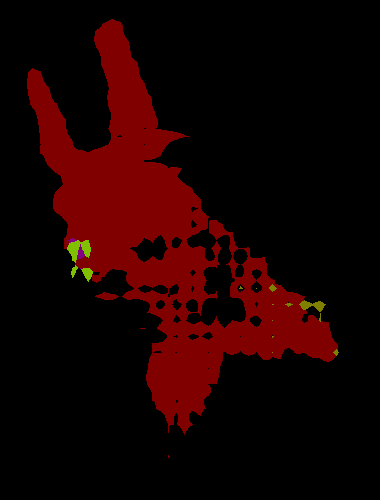}}~~
		\subfigure{\includegraphics[width=18mm]{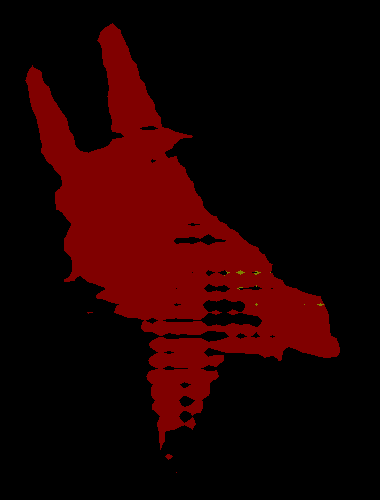}}~~
		\subfigure{\includegraphics[width=18mm]{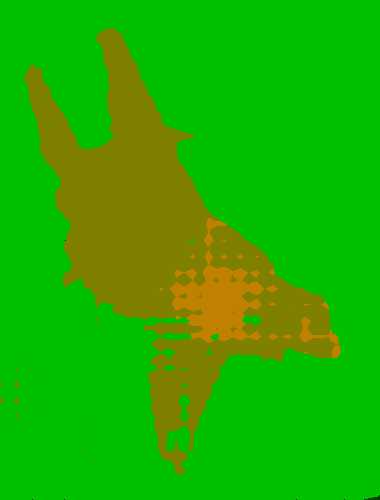}}~~\\
		\subfigure{\includegraphics[width=18mm]{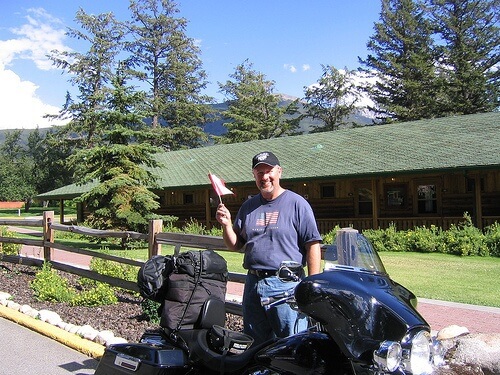}}~~
		\subfigure{\includegraphics[width=18mm]{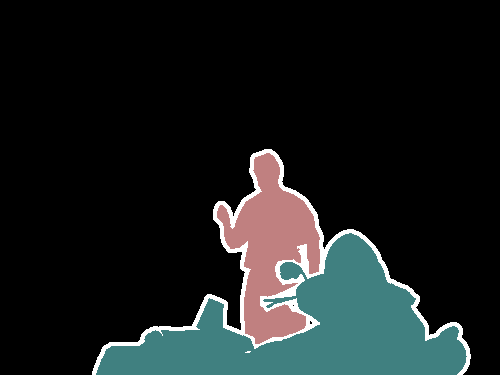}}~~
		\subfigure{\includegraphics[width=18mm]{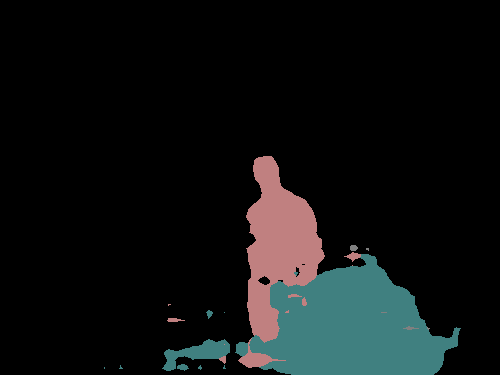}}~~
		\subfigure{\includegraphics[width=18mm]{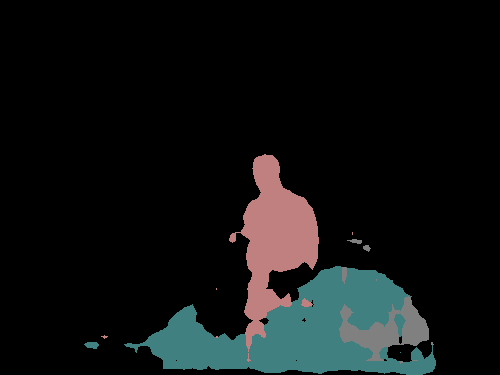}}~~
		\subfigure{\includegraphics[width=18mm]{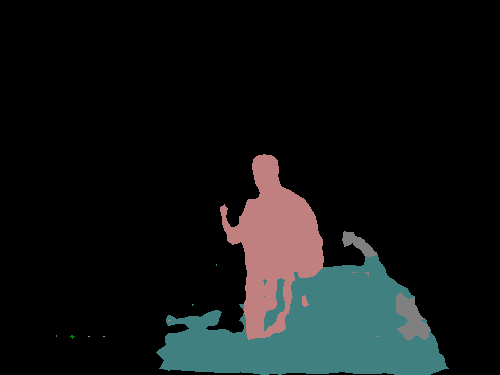}}~~
		\subfigure{\includegraphics[width=18mm]{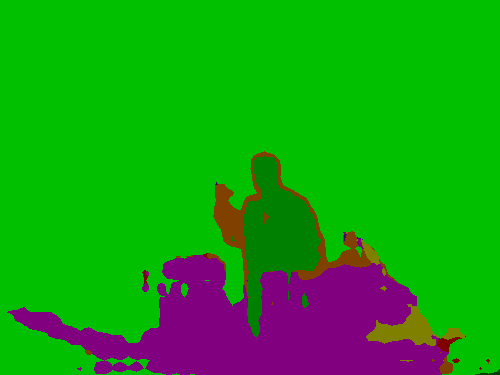}}~~\\
		\subfigure{\includegraphics[width=18mm]{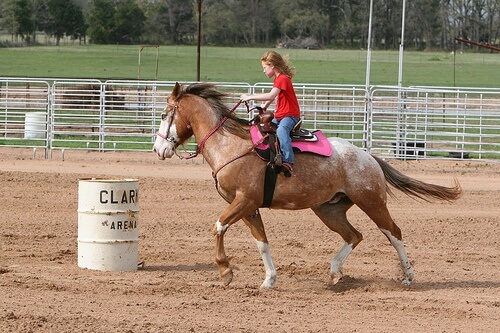}}~~
		\subfigure{\includegraphics[width=18mm]{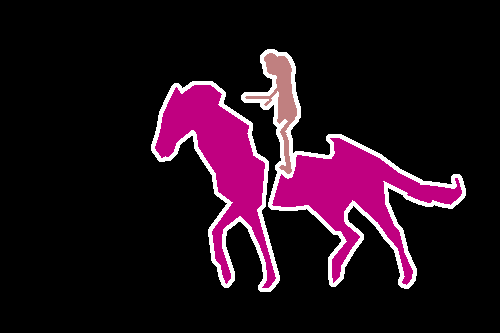}}~~
		\subfigure{\includegraphics[width=18mm]{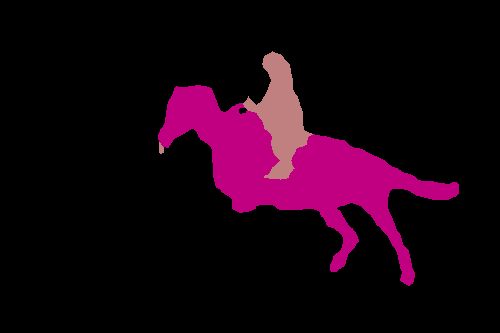}}~~
		\subfigure{\includegraphics[width=18mm]{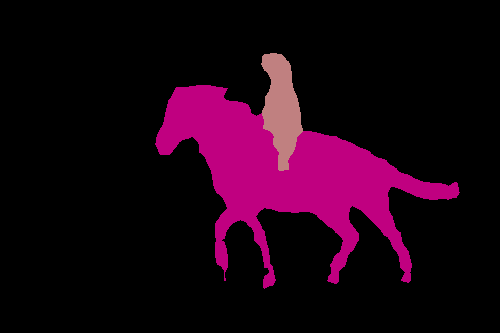}}~~
		\subfigure{\includegraphics[width=18mm]{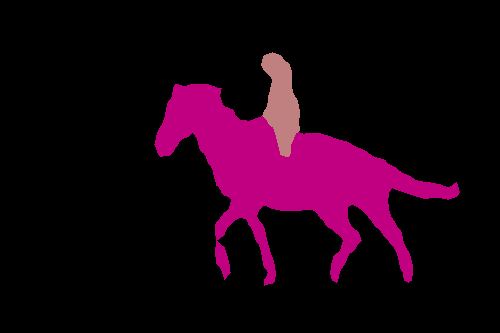}}~~
		\subfigure{\includegraphics[width=18mm]{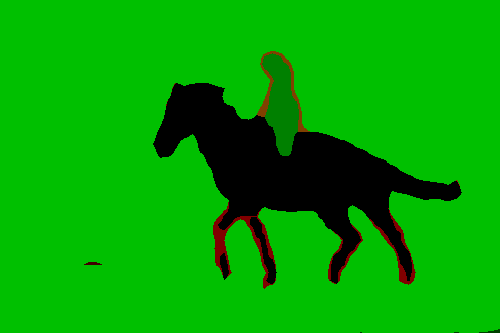}}~~\\
		\caption{Qualitative examples from the Pascal VOC 2012 val set. From left to right: image, ground truth, $L_{ce}$, proposed without adversarial loss, proposed, latent classes.    
		}
		\label{fig:qualitative_examples_VOC}
	\end{figure*}

\begin{figure*}
		\centering
		\subfigure{\includegraphics[width=25mm]{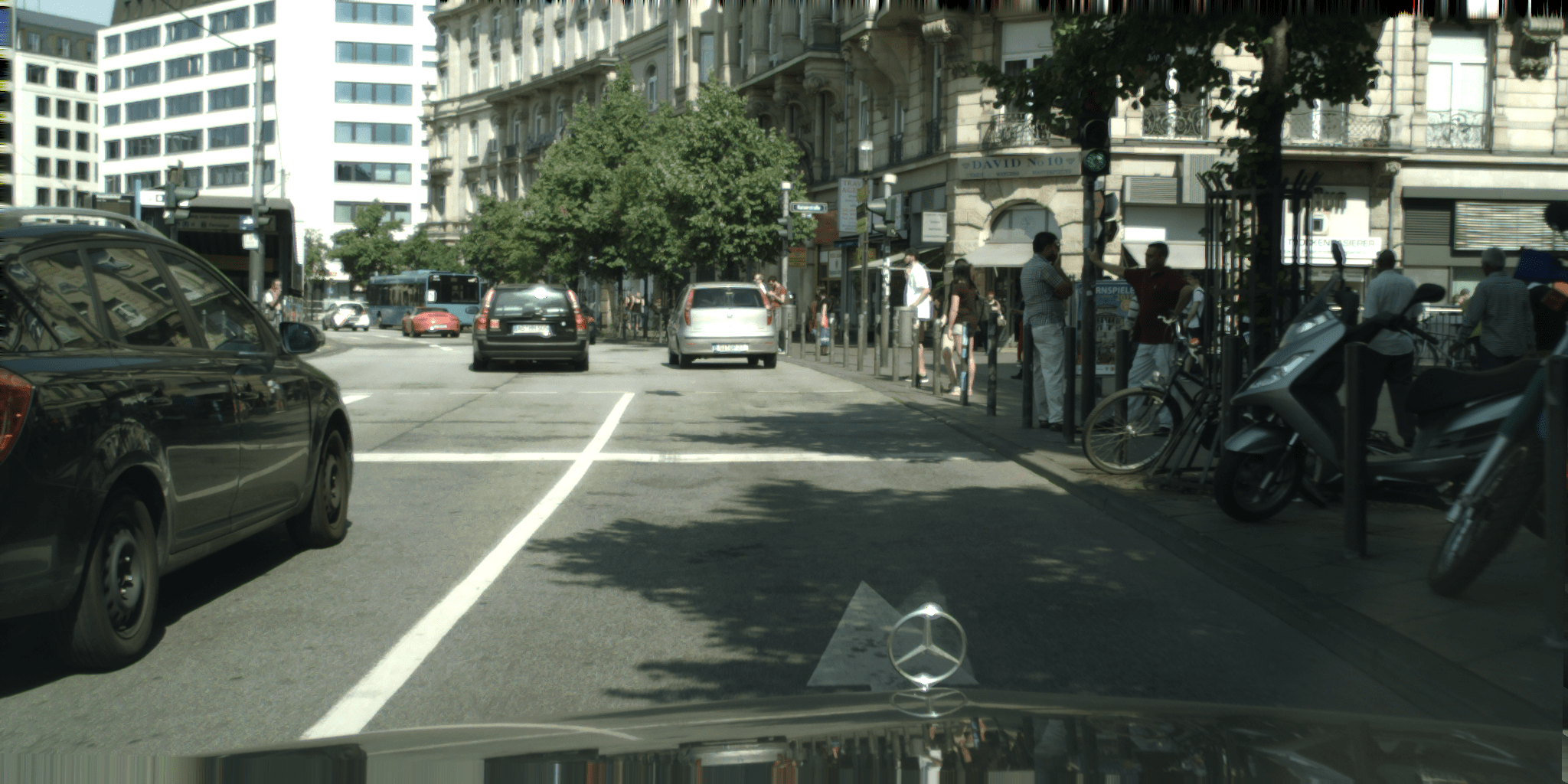}}~~
		\subfigure{\includegraphics[width=25mm]{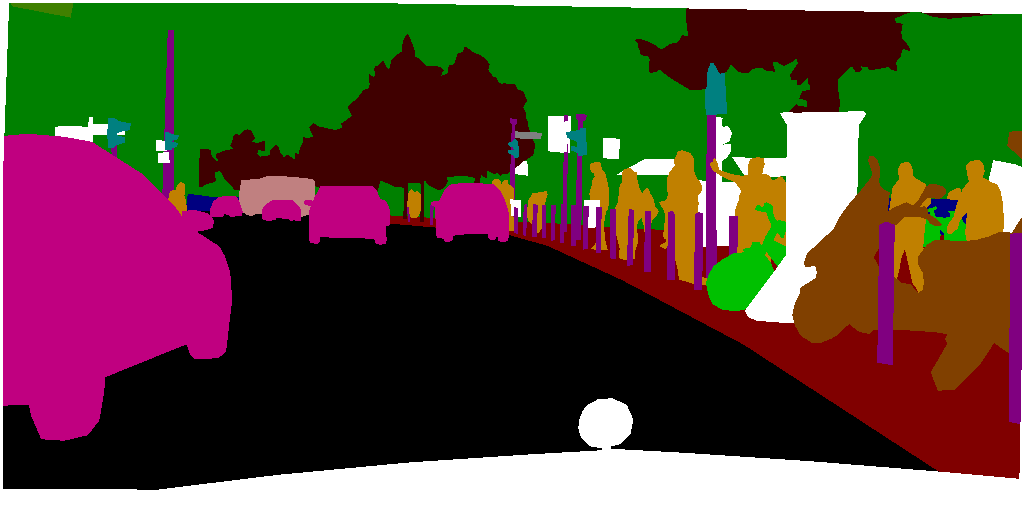}}~~
		\subfigure{\includegraphics[width=25mm]{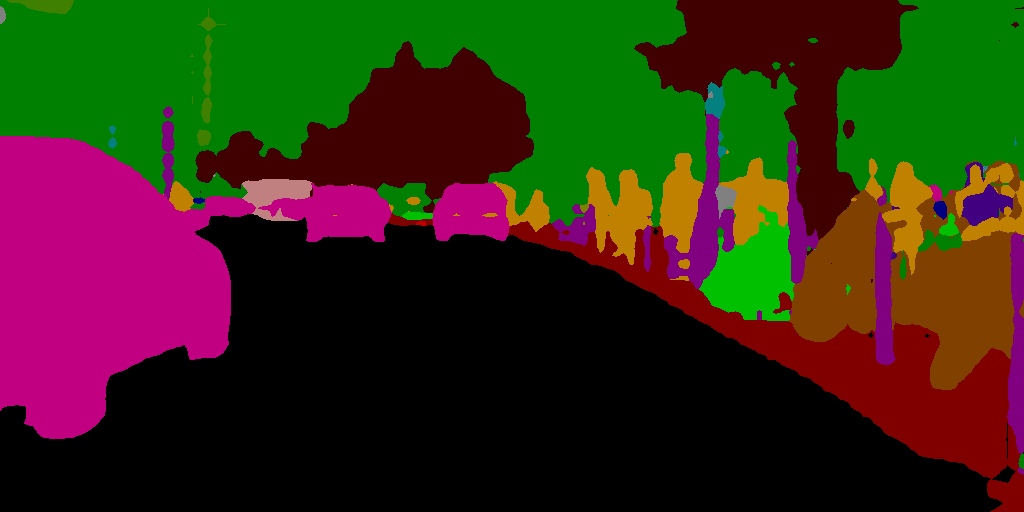}}~~
		\subfigure{\includegraphics[width=25mm]{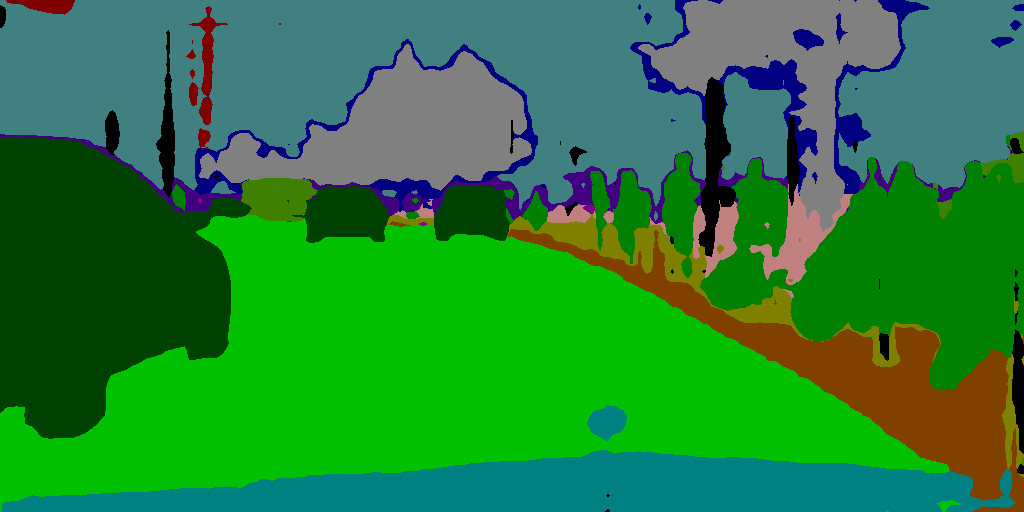}}~~\\
		\subfigure{\includegraphics[width=25mm]{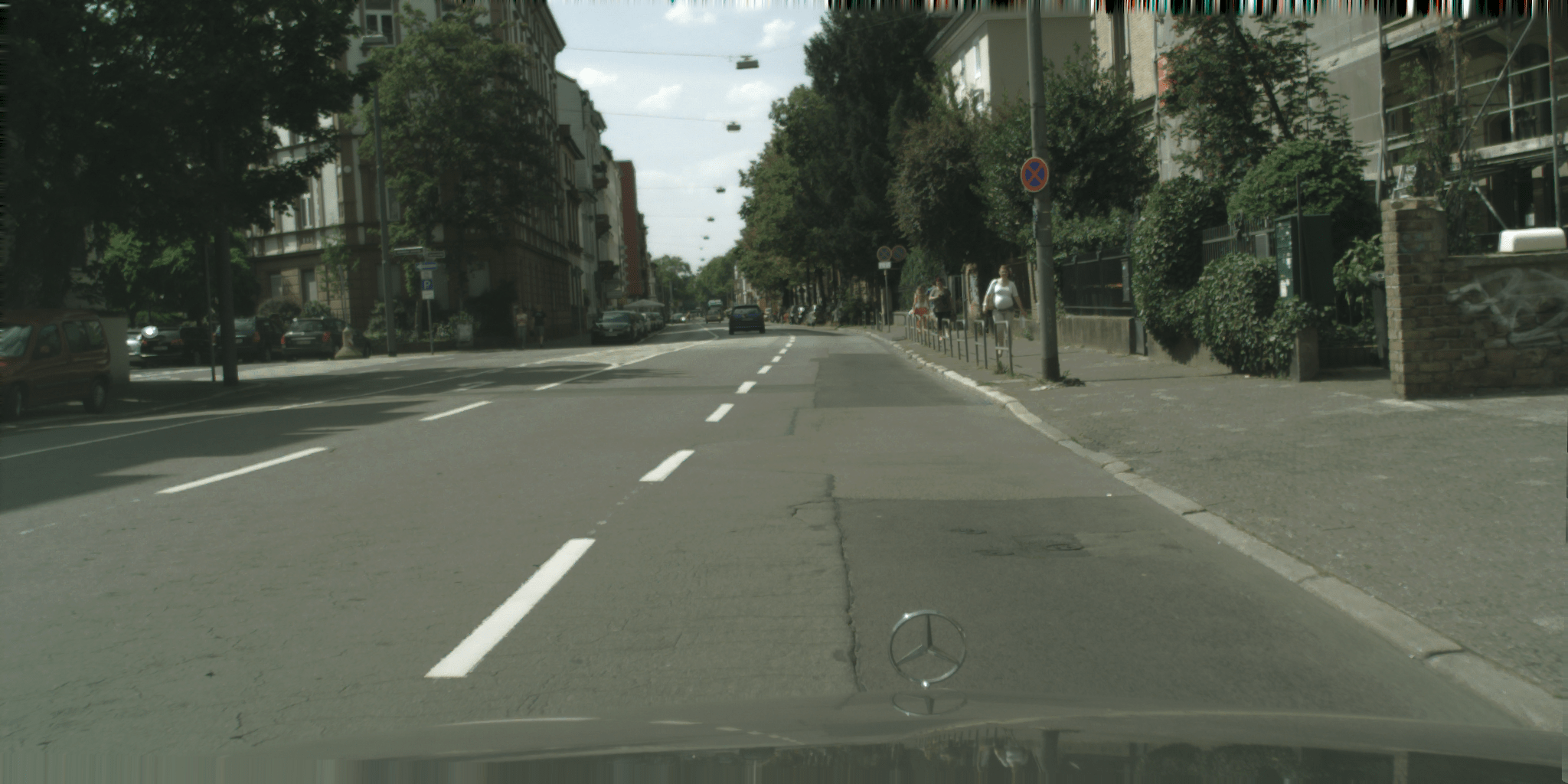}}~~
		\subfigure{\includegraphics[width=25mm]{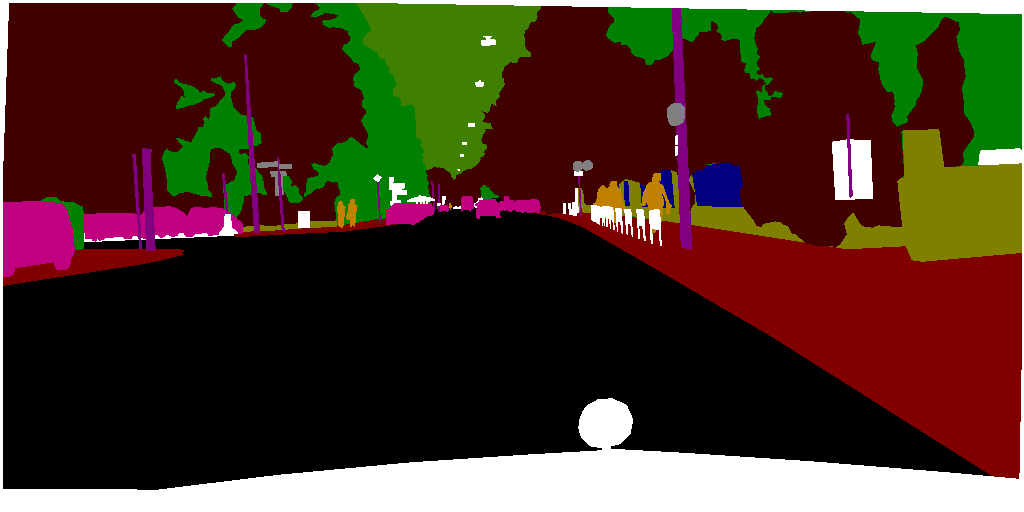}}~~
		\subfigure{\includegraphics[width=25mm]{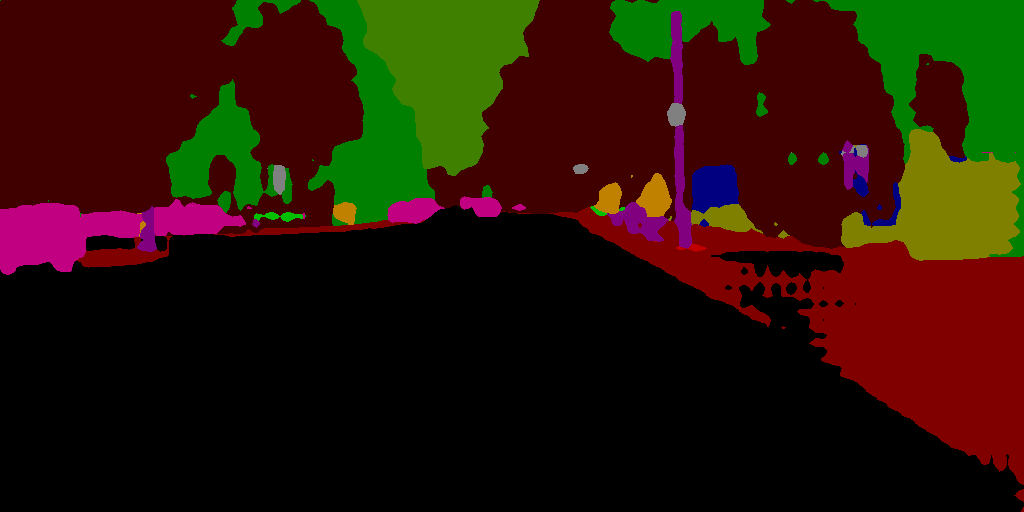}}~~
		\subfigure{\includegraphics[width=25mm]{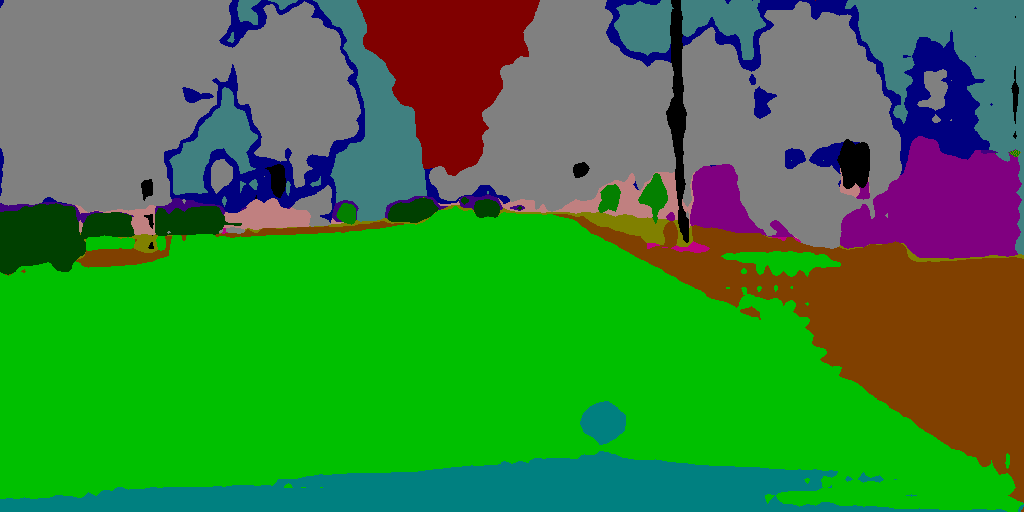}}~~\\
		\subfigure{\includegraphics[width=25mm]{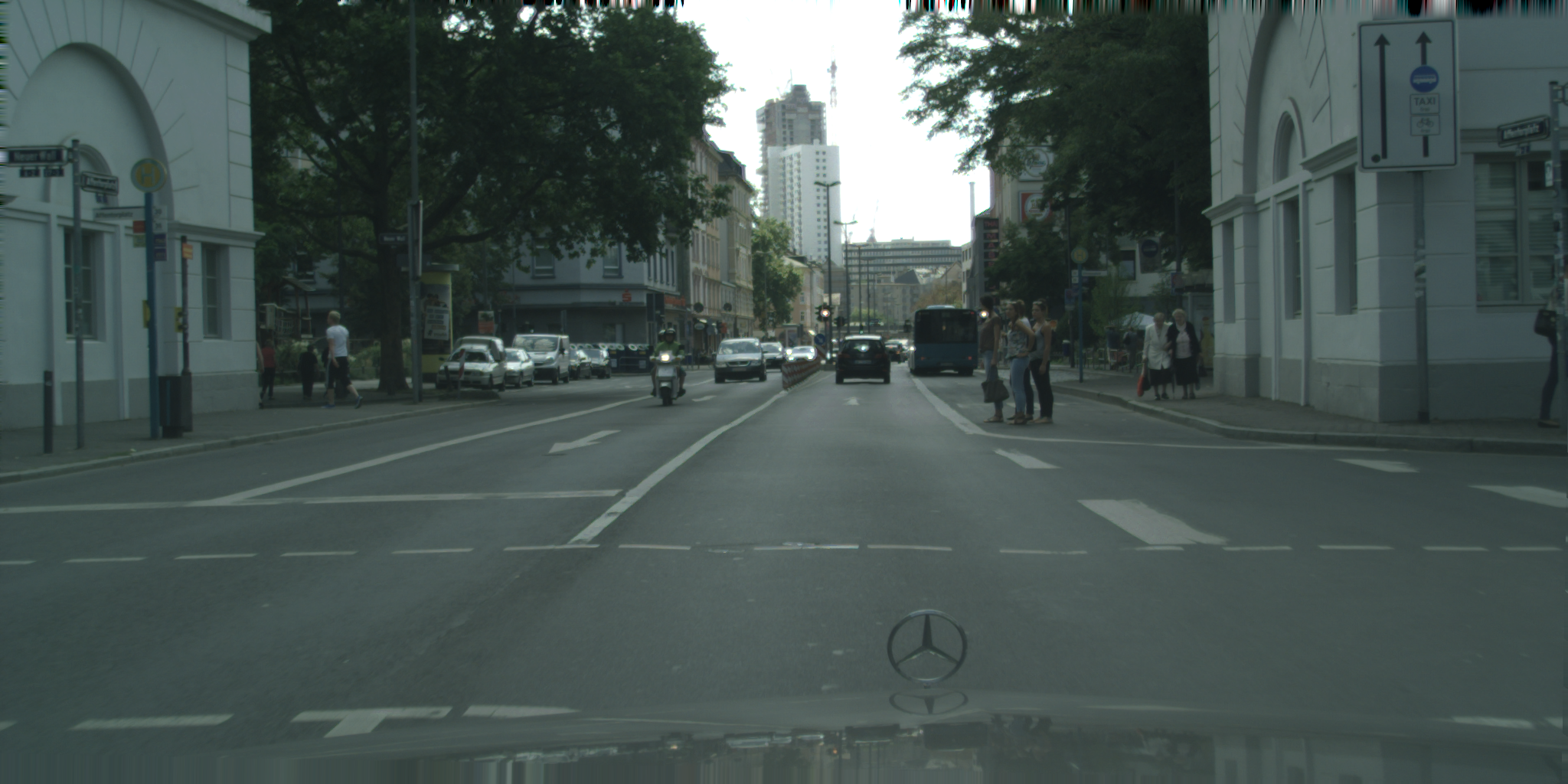}}~~
		\subfigure{\includegraphics[width=25mm]{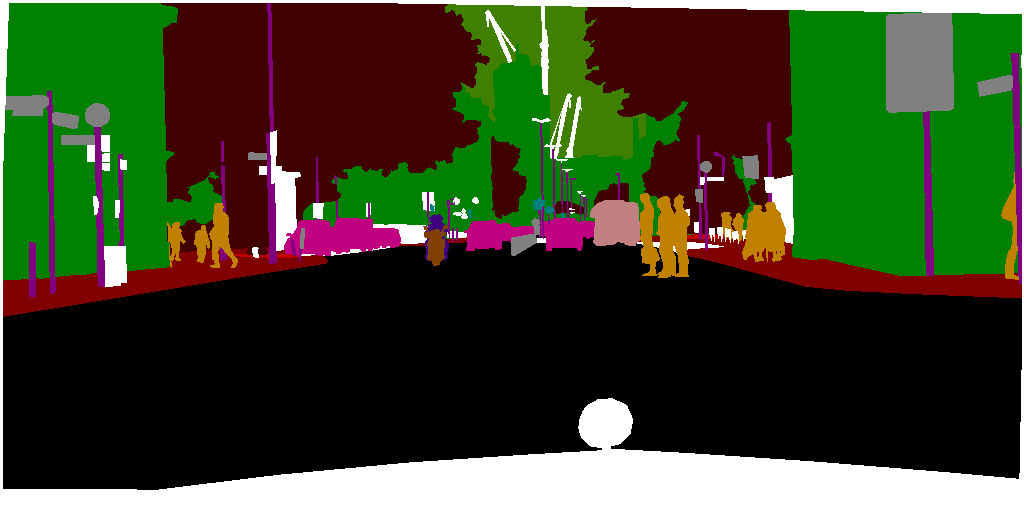}}~~
		\subfigure{\includegraphics[width=25mm]{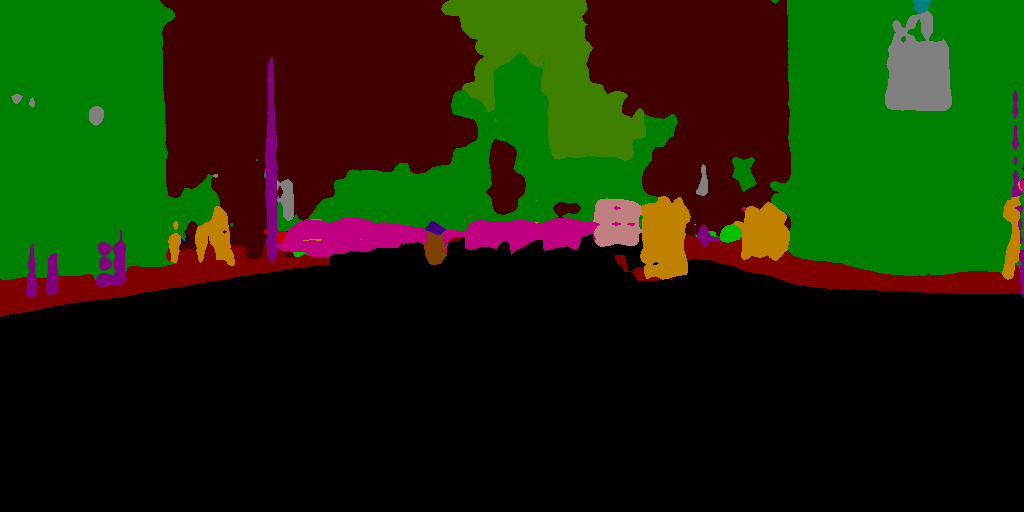}}~~
		\subfigure{\includegraphics[width=25mm]{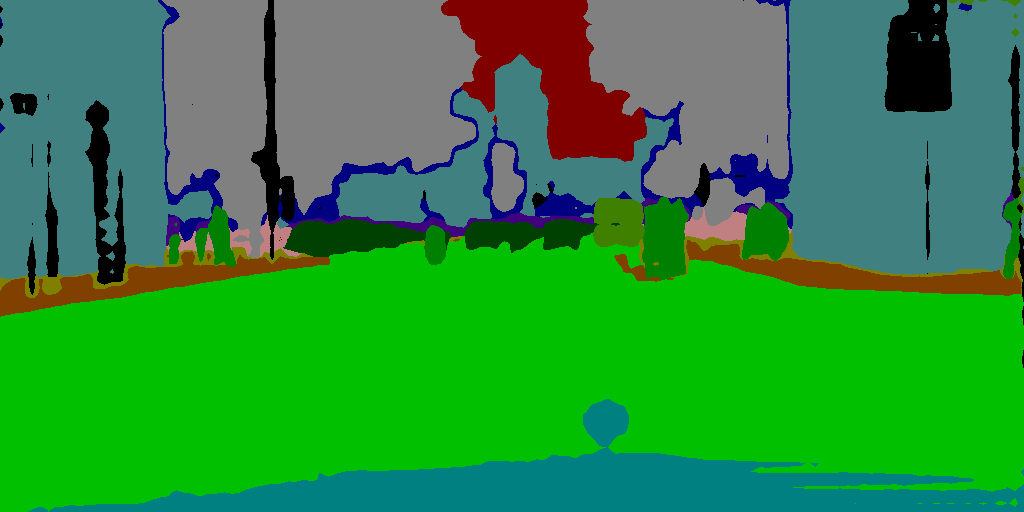}}~~\\
		\subfigure{\includegraphics[width=25mm]{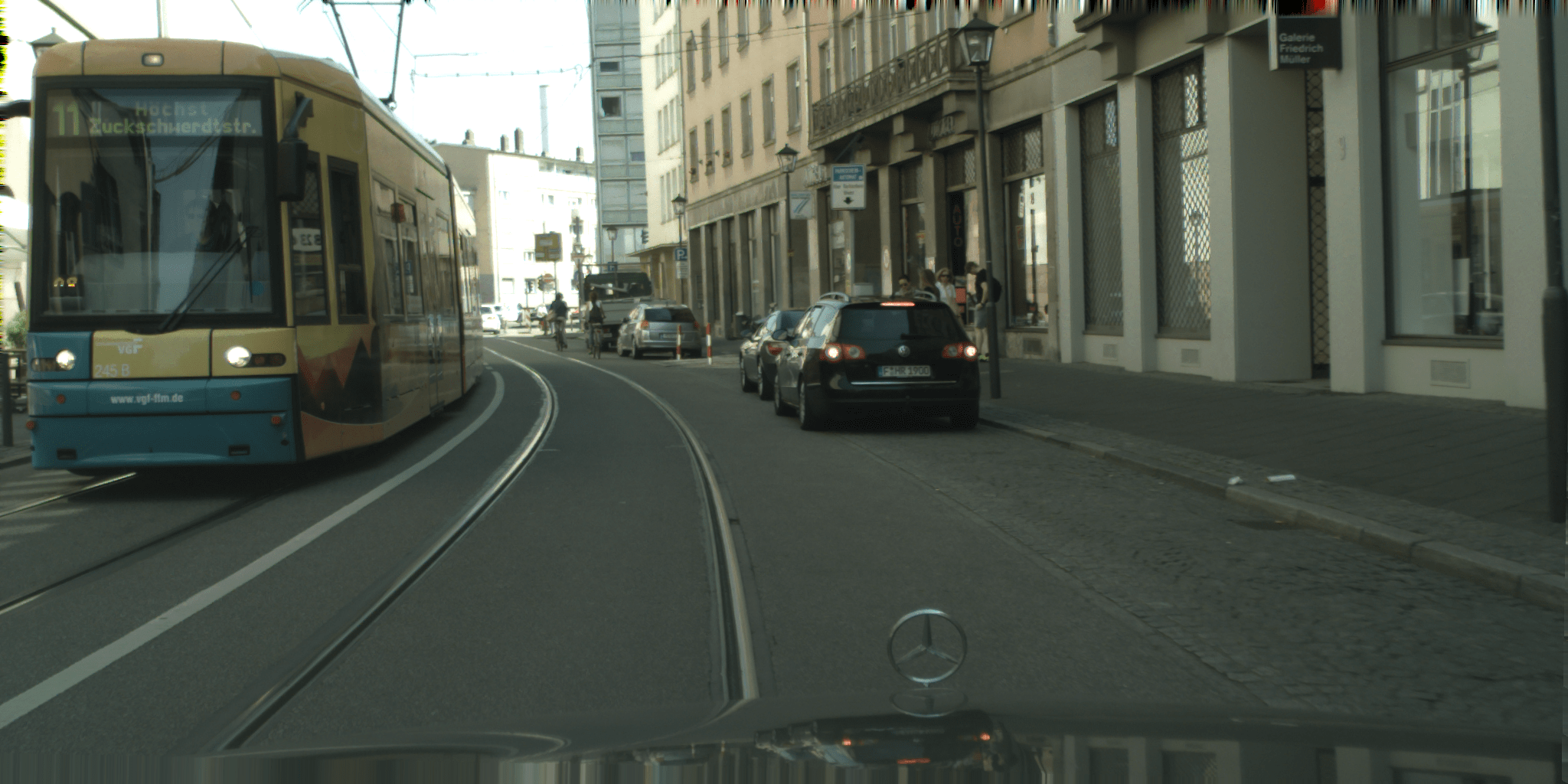}}~~
		\subfigure{\includegraphics[width=25mm]{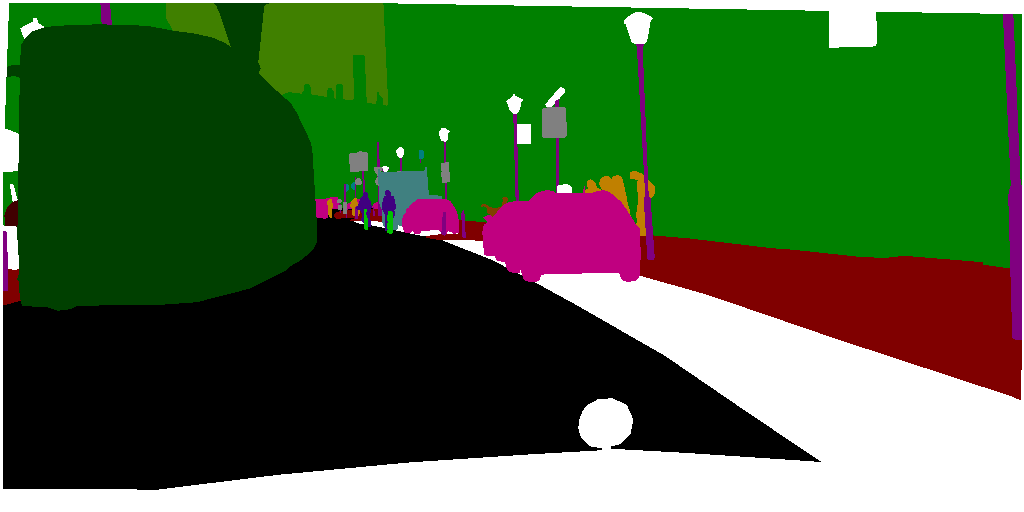}}~~
		\subfigure{\includegraphics[width=25mm]{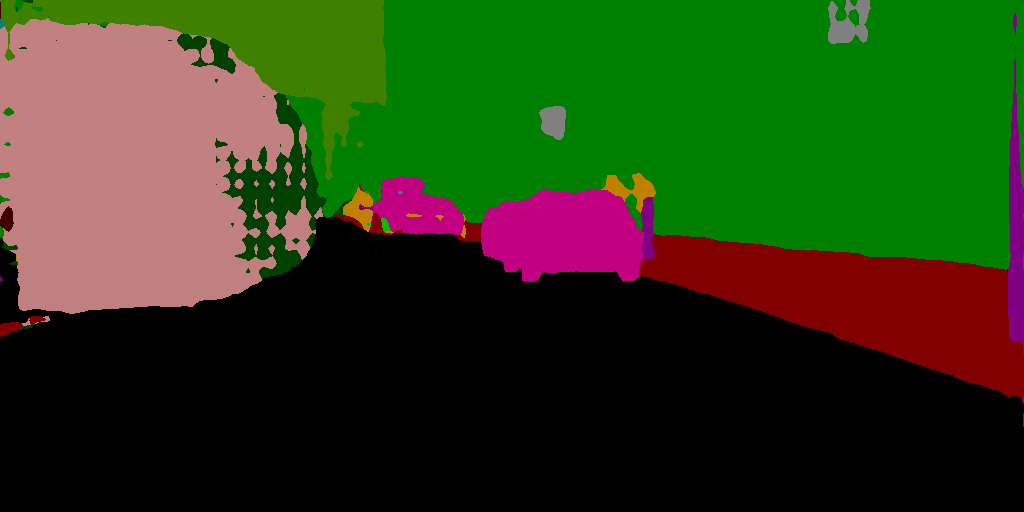}}~~
		\subfigure{\includegraphics[width=25mm]{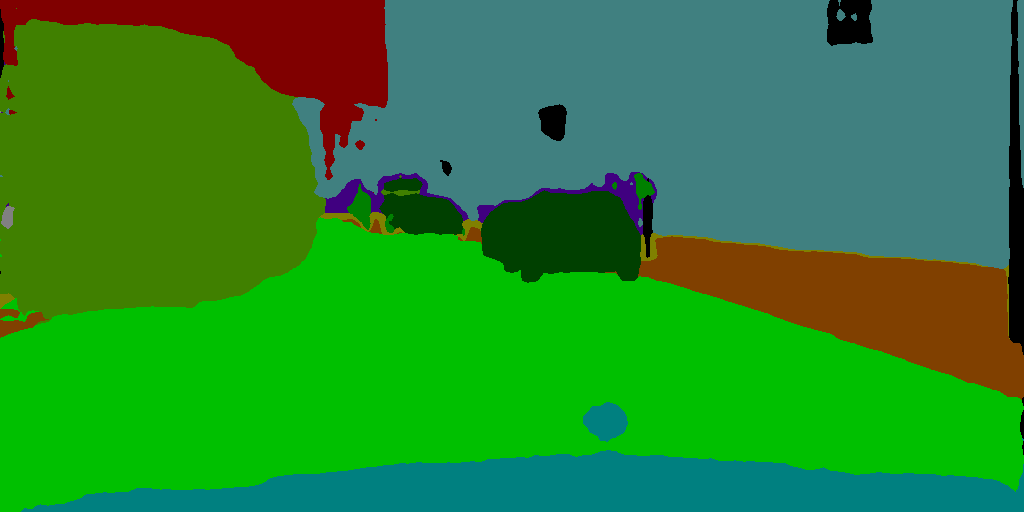}}~~\\
		\subfigure{\includegraphics[width=25mm]{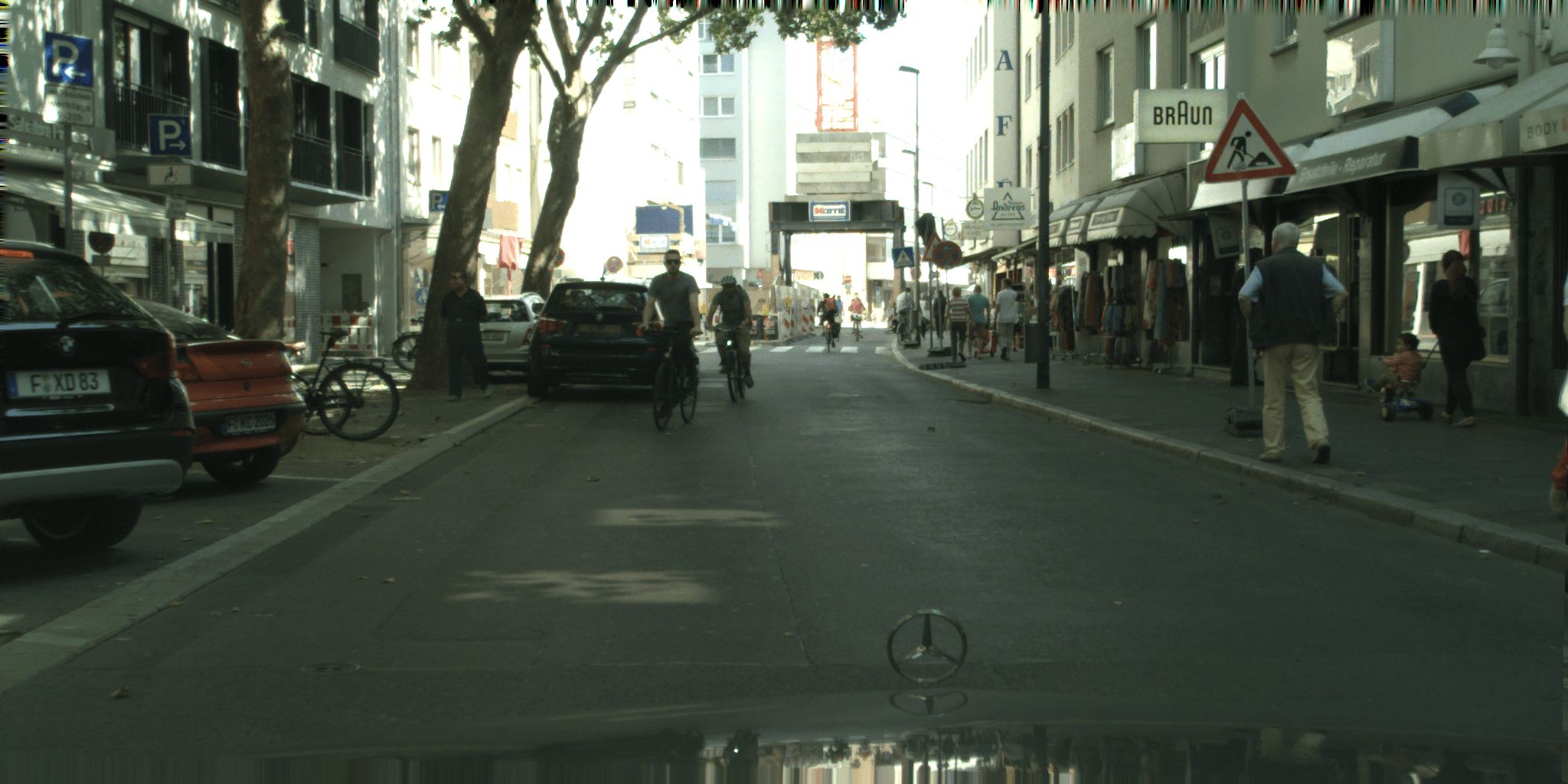}}~~
		\subfigure{\includegraphics[width=25mm]{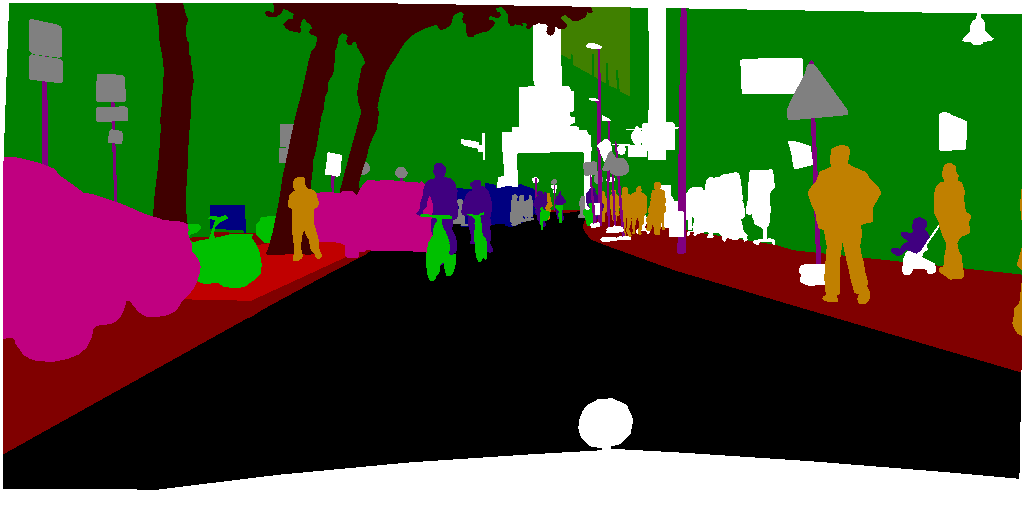}}~~
		\subfigure{\includegraphics[width=25mm]{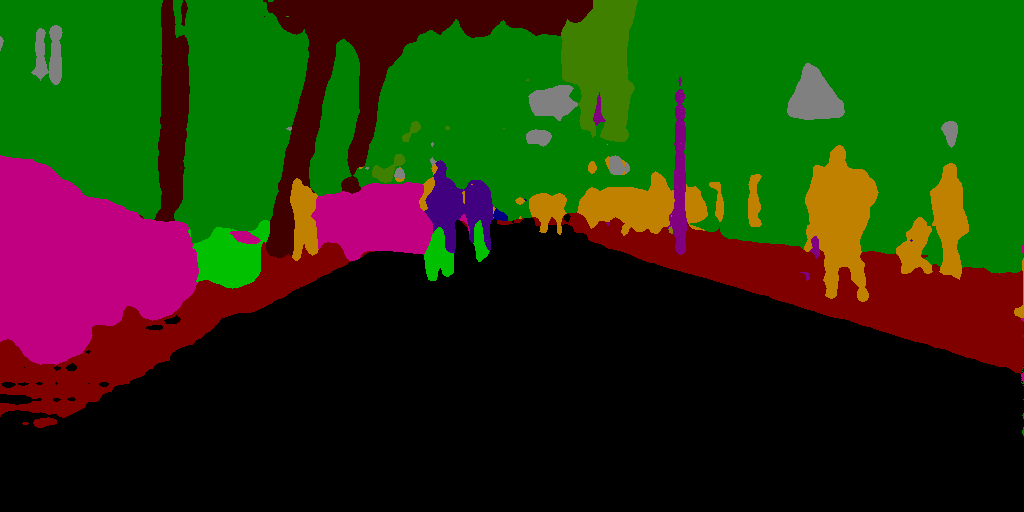}}~~
		\subfigure{\includegraphics[width=25mm]{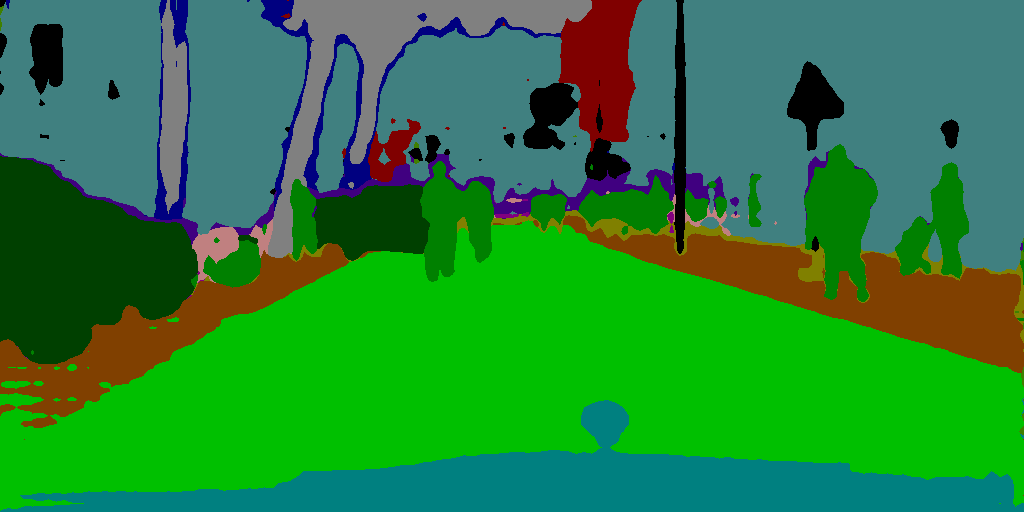}}~~\\
		\subfigure{\includegraphics[width=25mm]{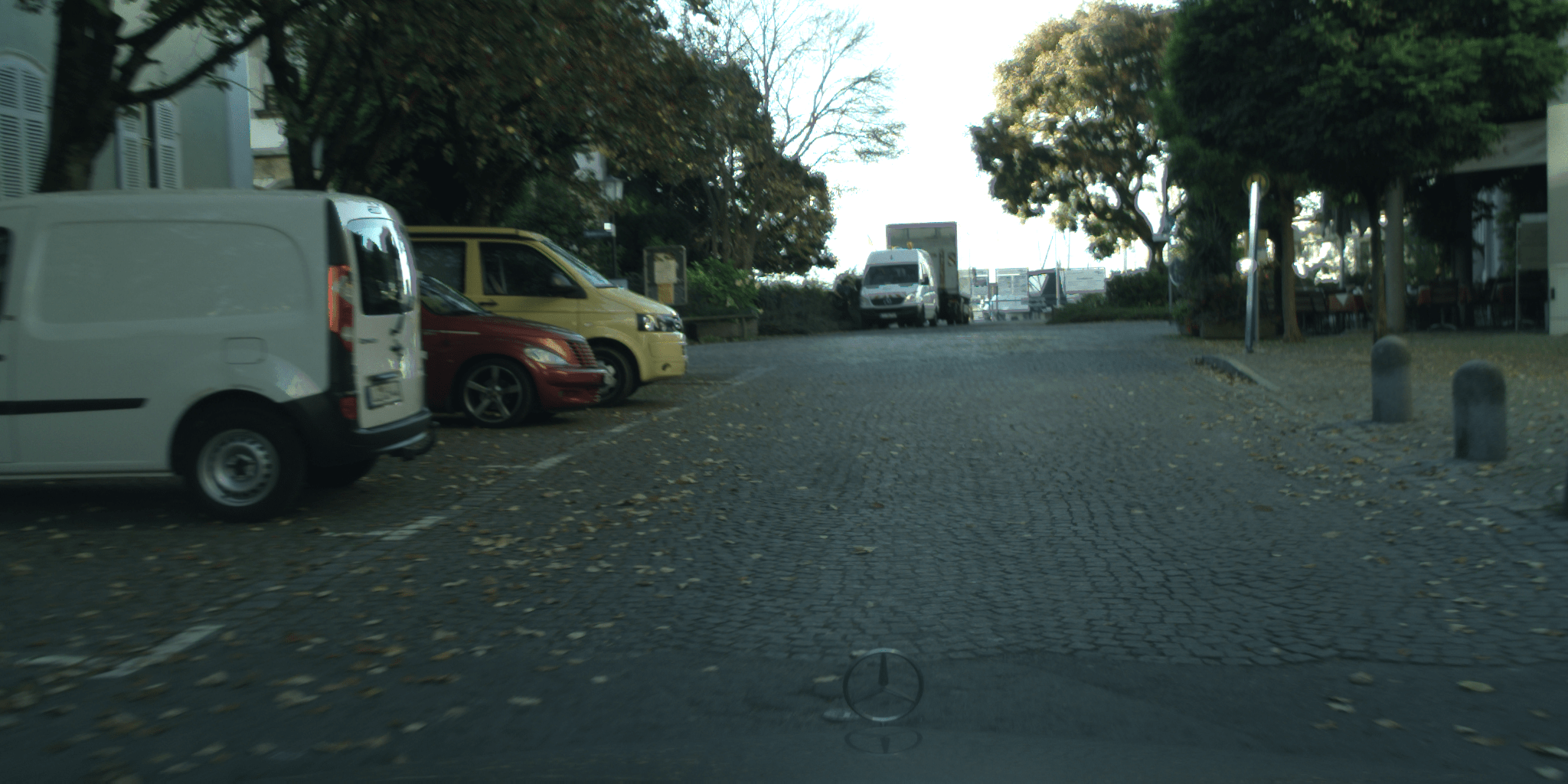}}~~
		\subfigure{\includegraphics[width=25mm]{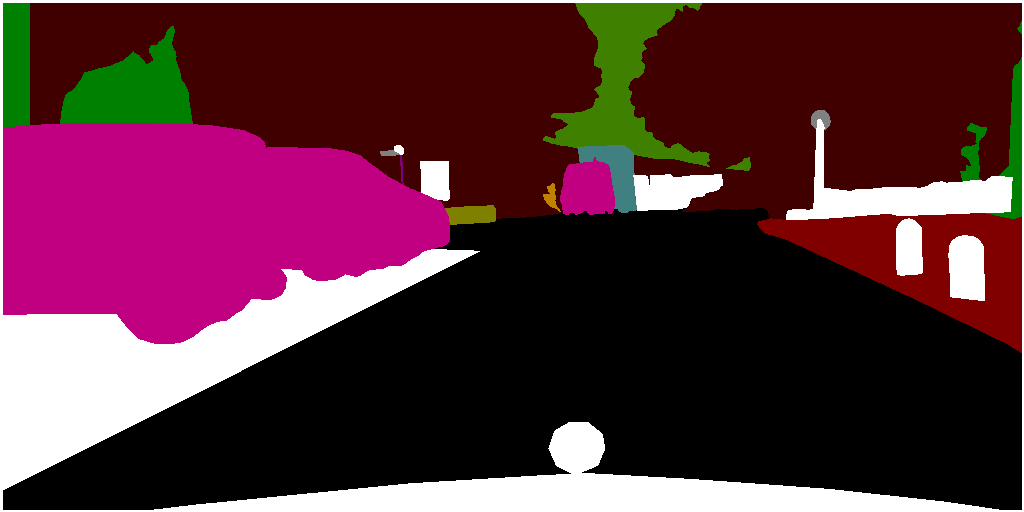}}~~
		\subfigure{\includegraphics[width=25mm]{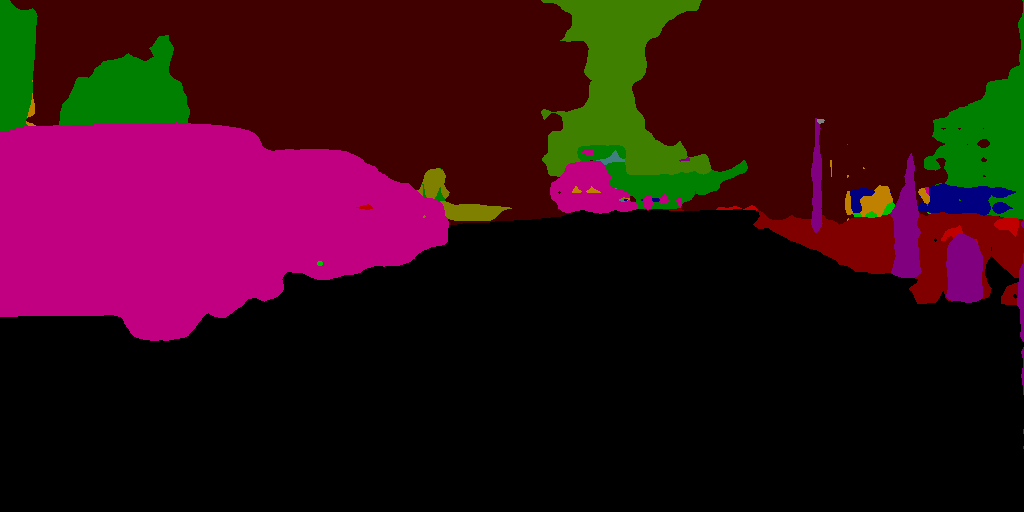}}~~
		\subfigure{\includegraphics[width=25mm]{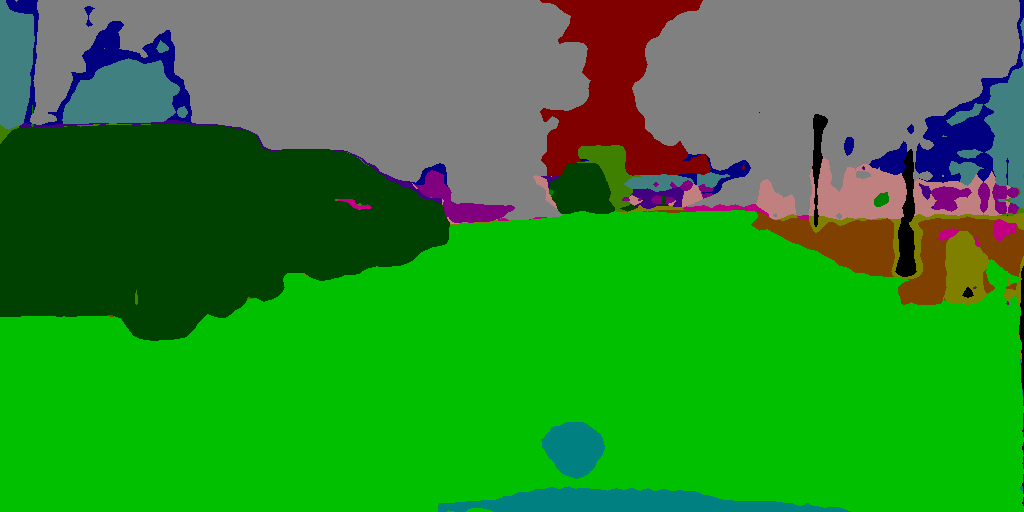}}~~\\
		\subfigure{\includegraphics[width=25mm]{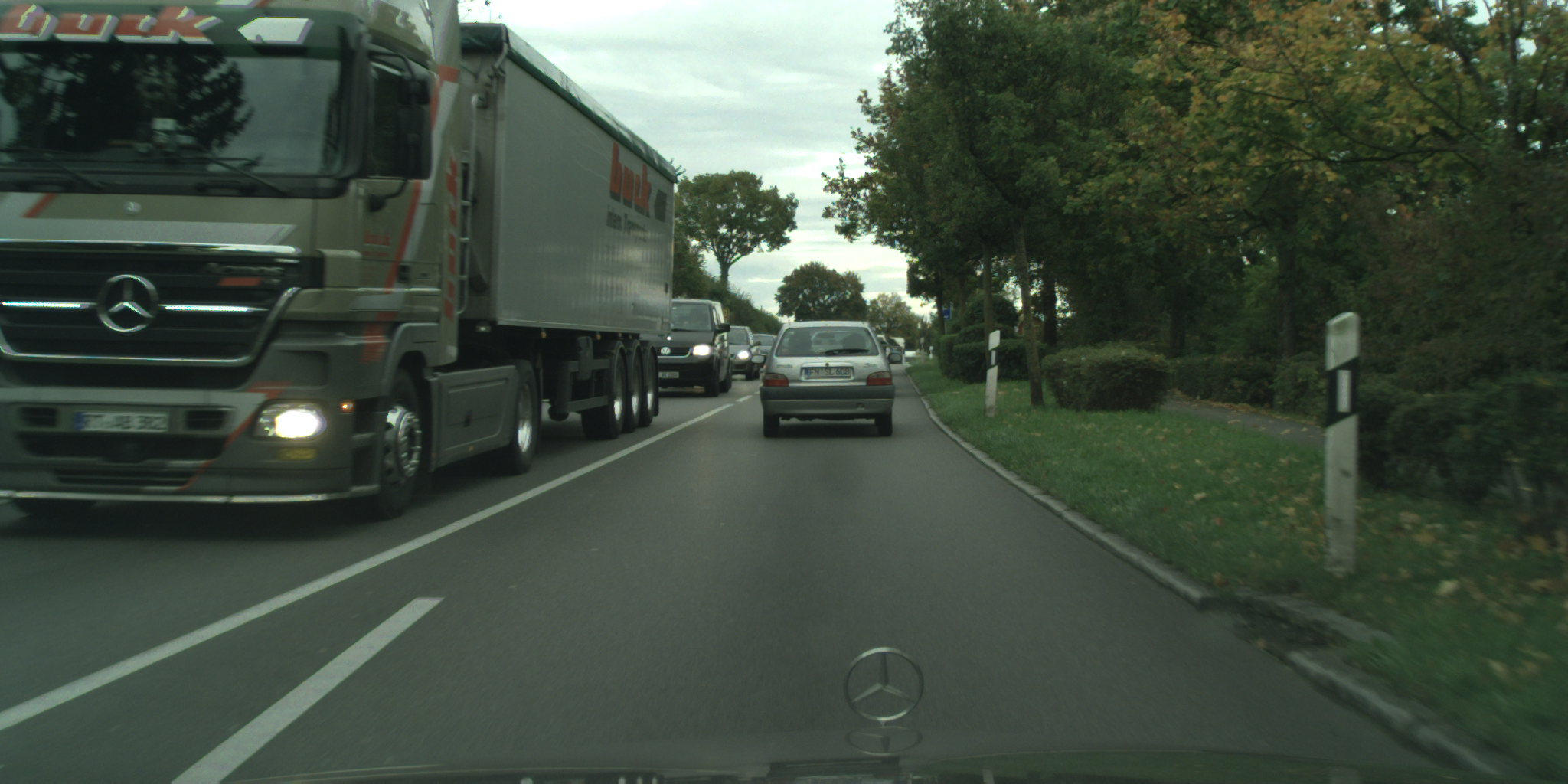}}~~
		\subfigure{\includegraphics[width=25mm]{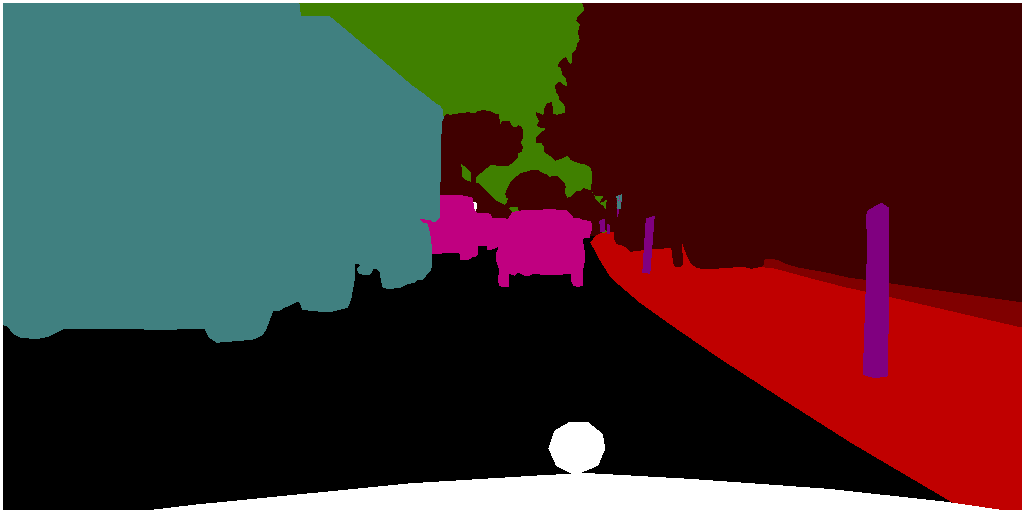}}~~
		\subfigure{\includegraphics[width=25mm]{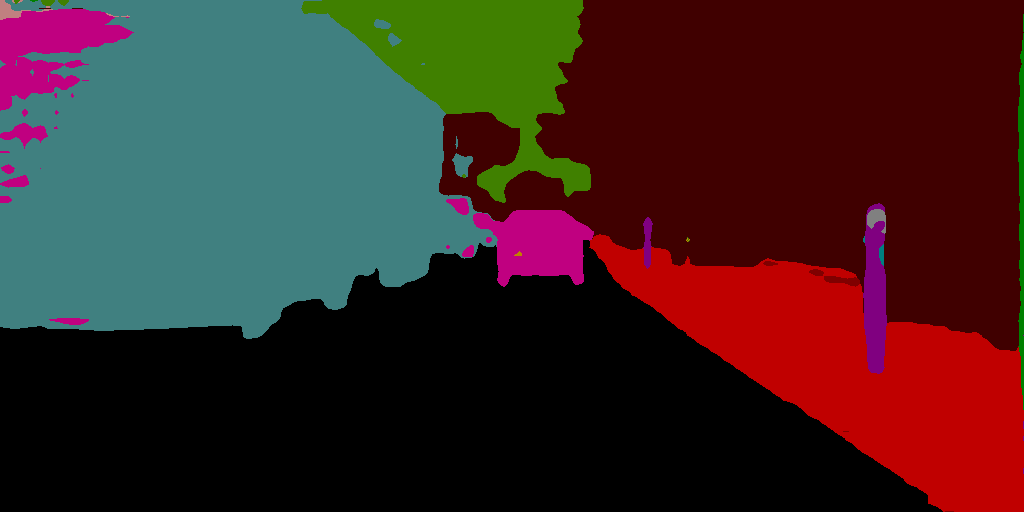}}~~
		\subfigure{\includegraphics[width=25mm]{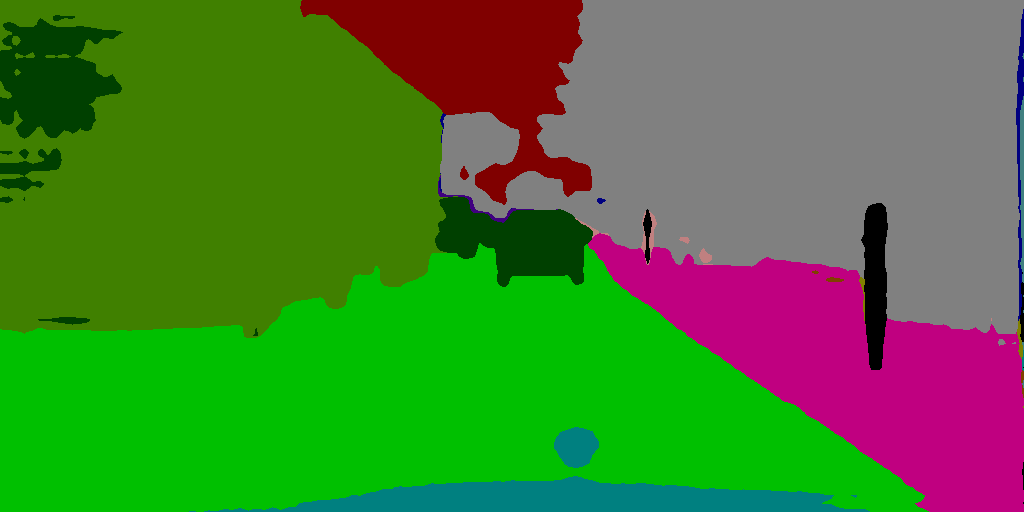}}~~\\
		\subfigure{\includegraphics[width=25mm]{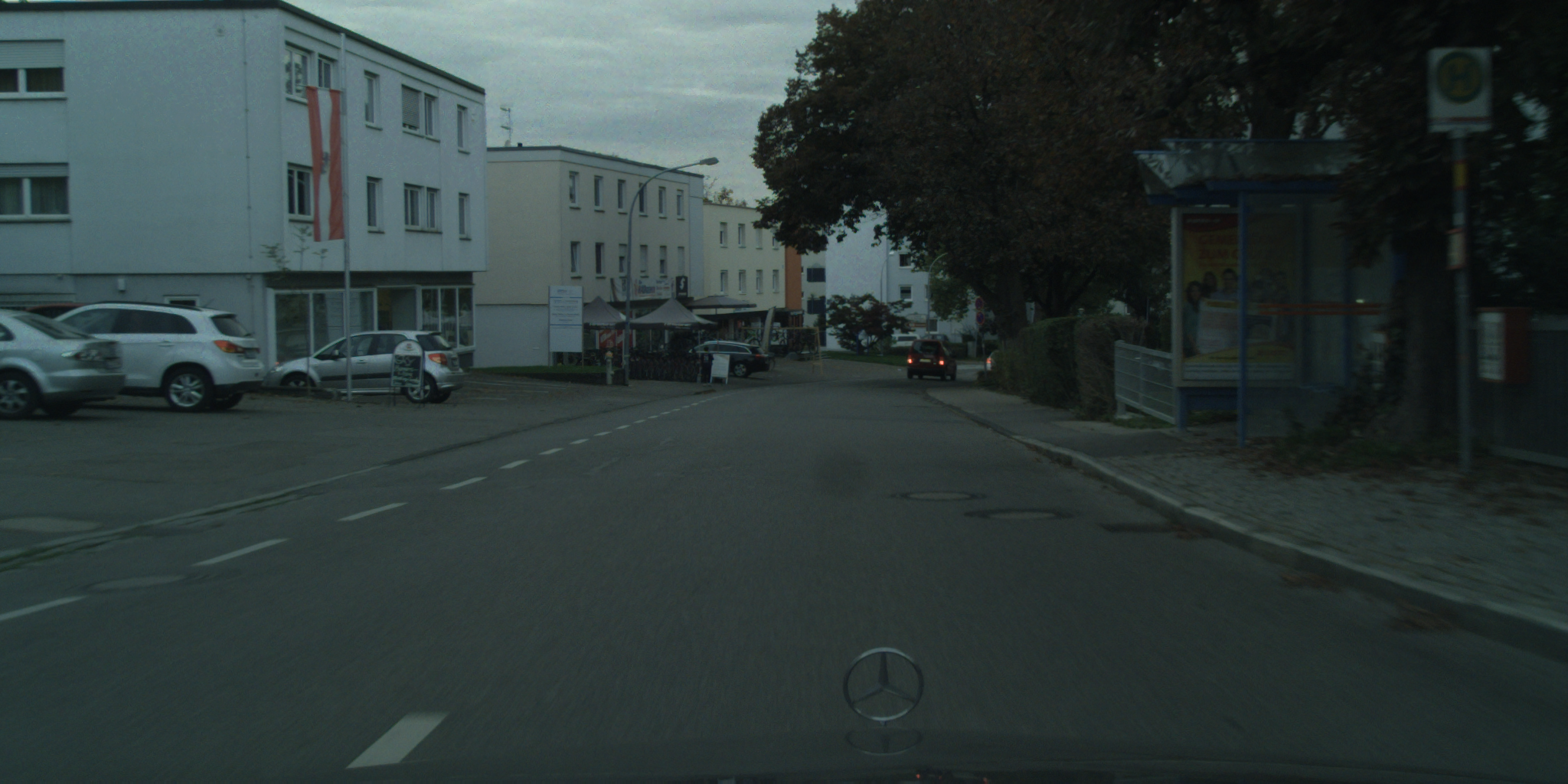}}~~
		\subfigure{\includegraphics[width=25mm]{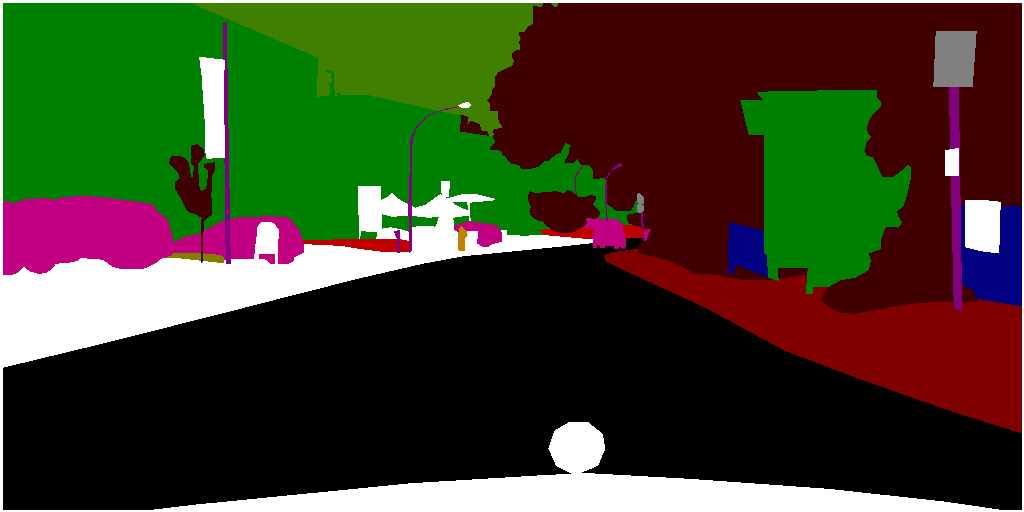}}~~
		\subfigure{\includegraphics[width=25mm]{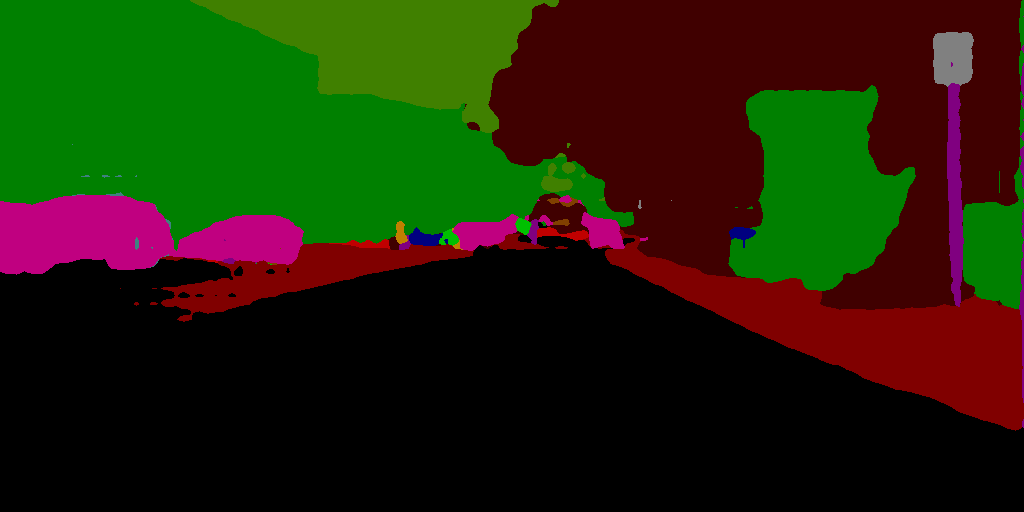}}~~
		\subfigure{\includegraphics[width=25mm]{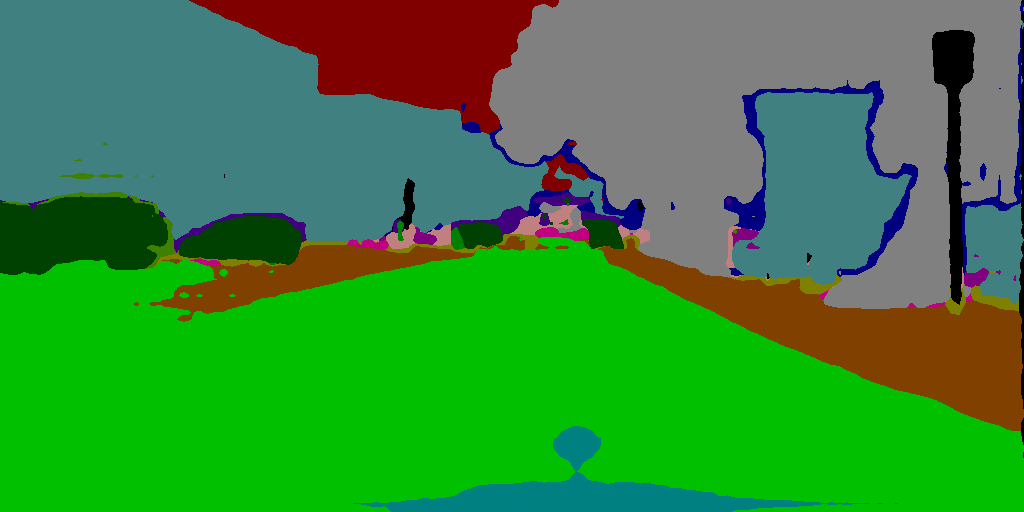}}~~\\
		\subfigure{\includegraphics[width=25mm]{examples_in_paper20/munster_000033_000019_leftImg8bit.png}}~~
		\subfigure{\includegraphics[width=25mm]{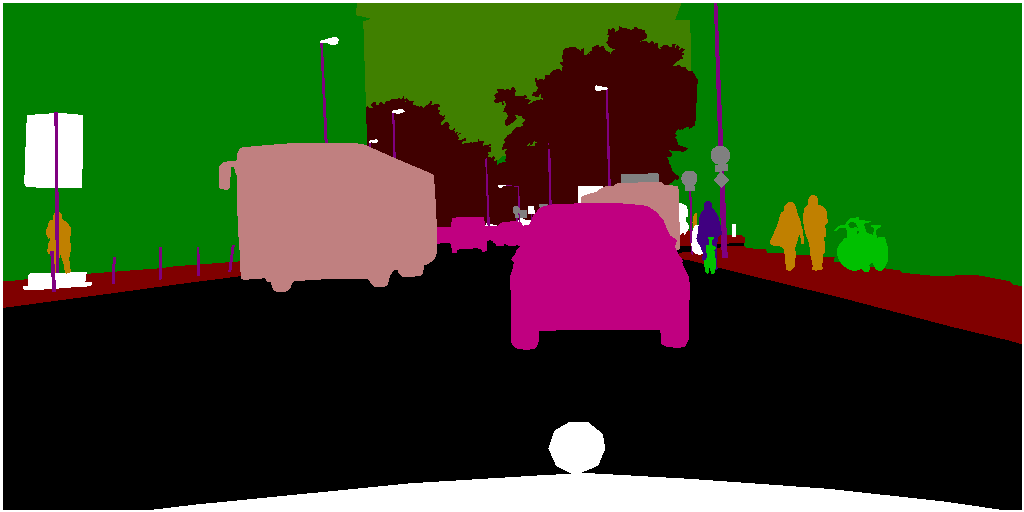}}~~
		\subfigure{\includegraphics[width=25mm]{examples_in_paper20/munster_000033_000019.png}}~~
		\subfigure{\includegraphics[width=25mm]{examples_in_paper20/munster_000033_000019_lv.png}}~~\\
		\subfigure{\includegraphics[width=25mm]{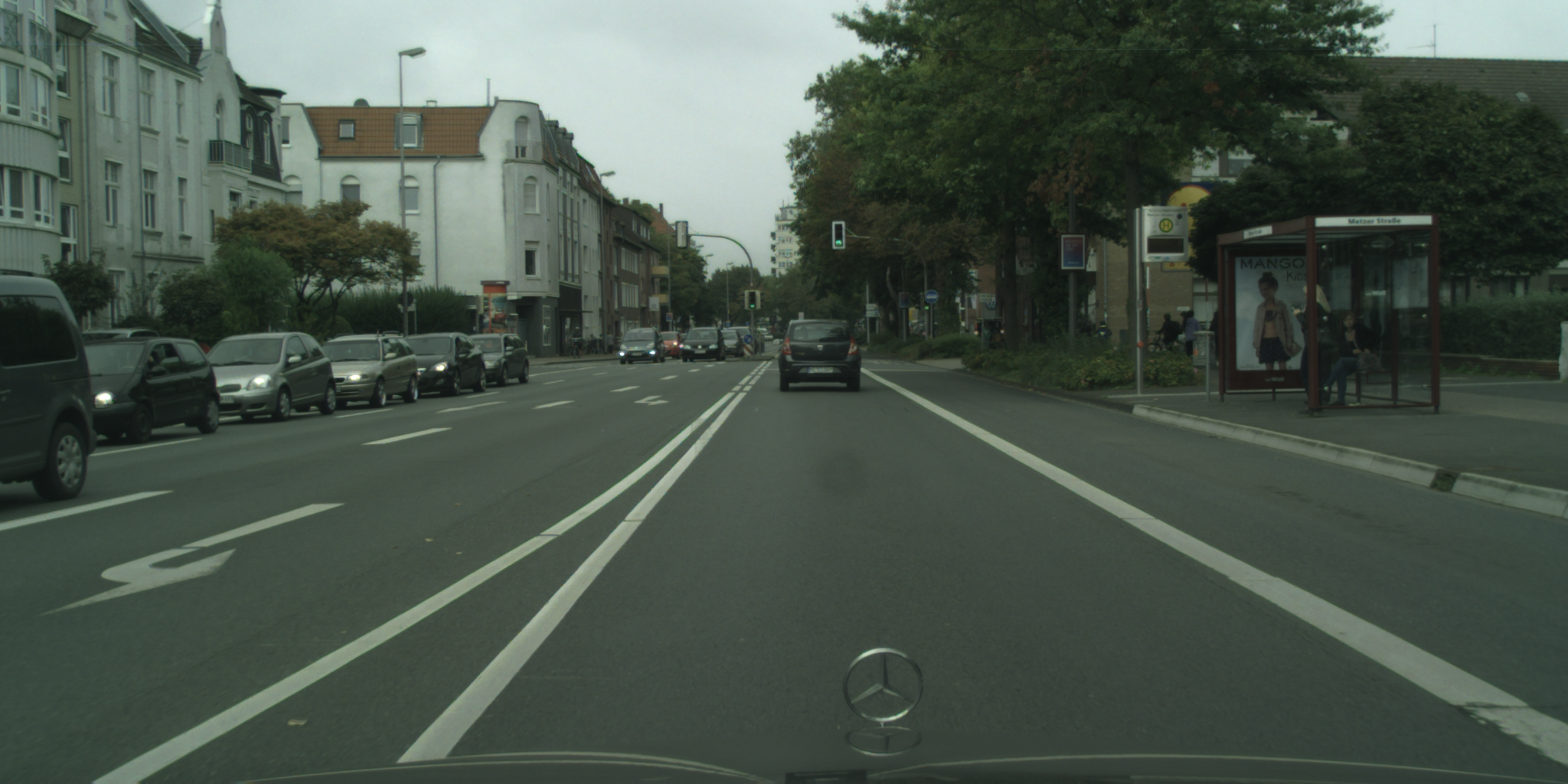}}~~
		\subfigure{\includegraphics[width=25mm]{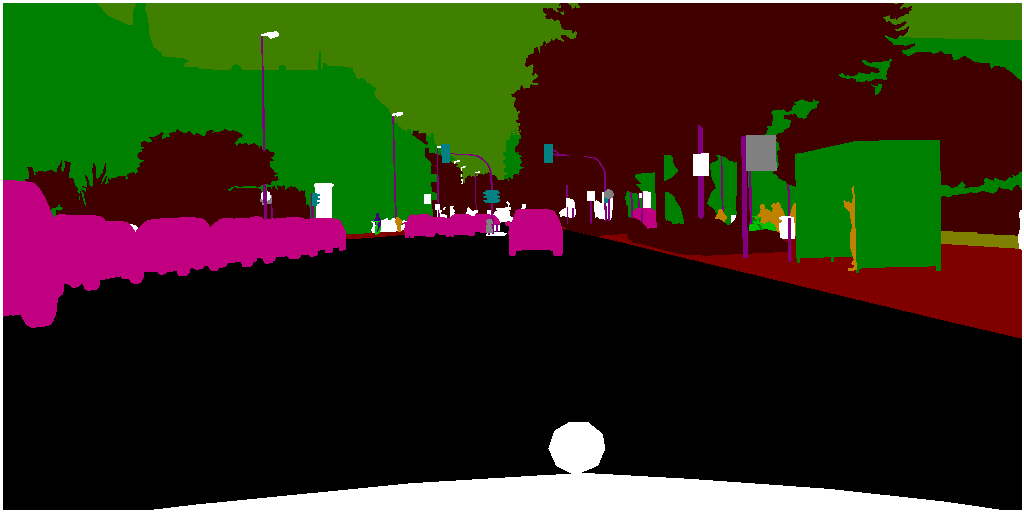}}~~
		\subfigure{\includegraphics[width=25mm]{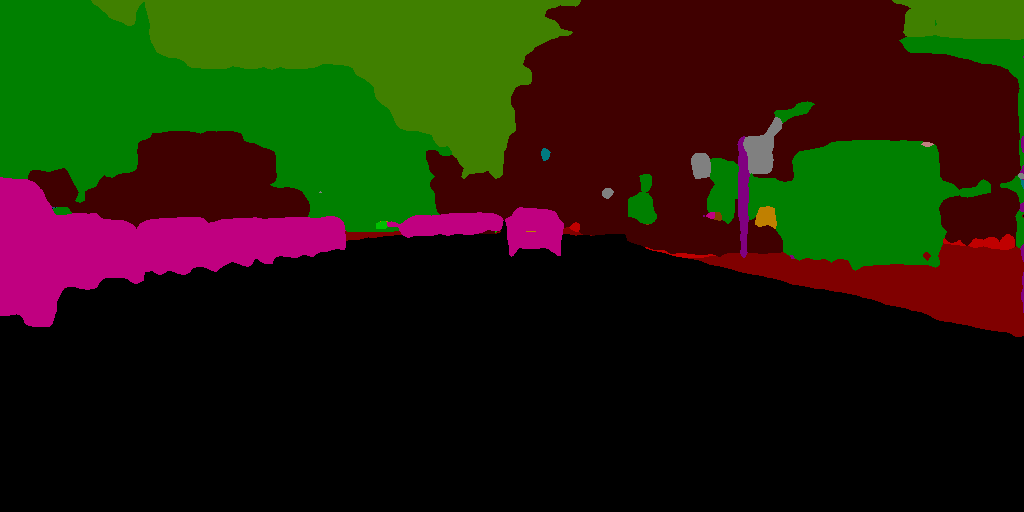}}~~
		\subfigure{\includegraphics[width=25mm]{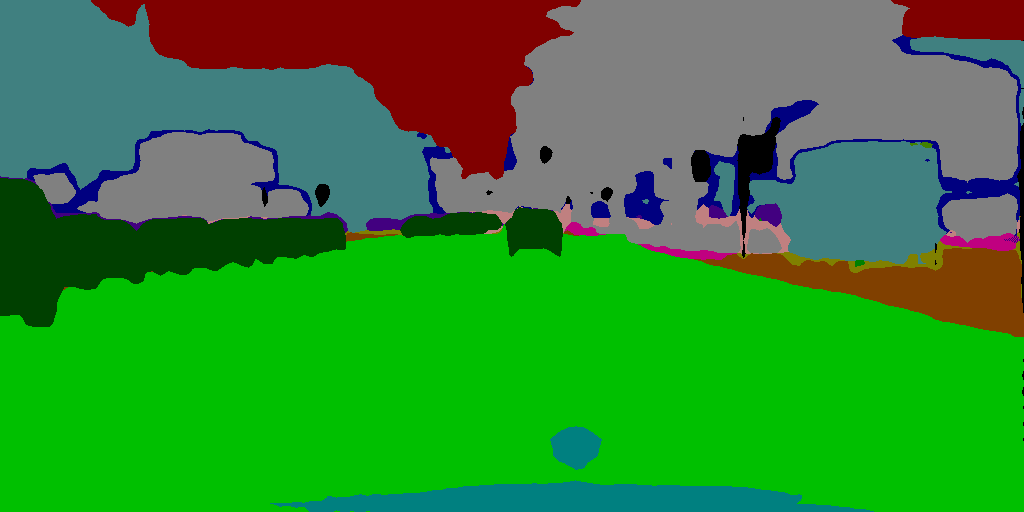}}~~\\
		\caption{Qualitative examples from the  Cityscapes val set. From left to right: image, ground truth, proposed, latent classes.}
		\label{fig:qualitative_examplesCS}
	\end{figure*}

\begin{figure*}
		\centering
		\subfigure{\includegraphics[width=25mm]{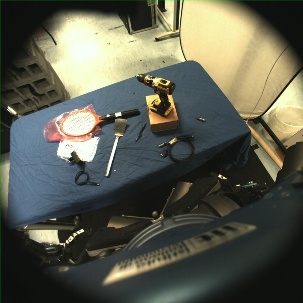}}~~
		\subfigure{\includegraphics[width=25mm]{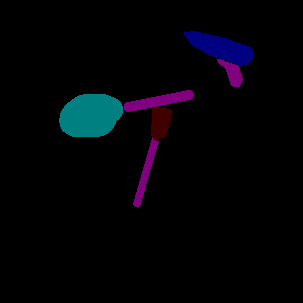}}~~
		\subfigure{\includegraphics[width=25mm]{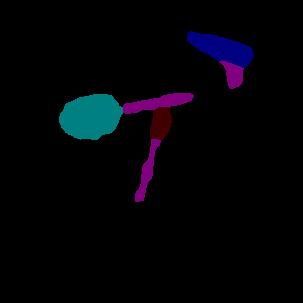}}~~
		\subfigure{\includegraphics[width=25mm]{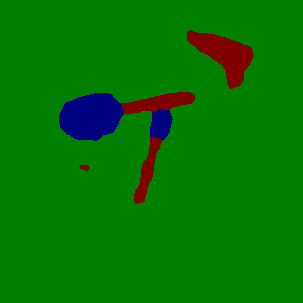}}~~\\
		\subfigure{\includegraphics[width=25mm]{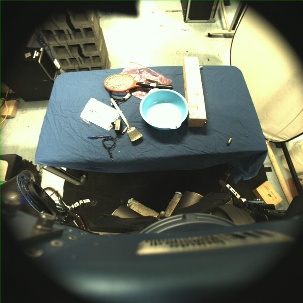}}~~
		\subfigure{\includegraphics[width=25mm]{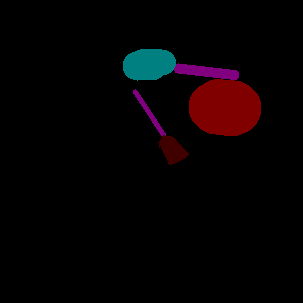}}~~
		\subfigure{\includegraphics[width=25mm]{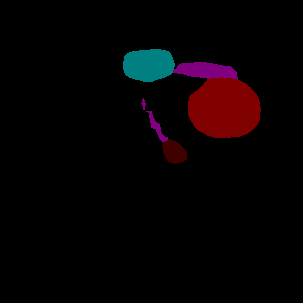}}~~
		\subfigure{\includegraphics[width=25mm]{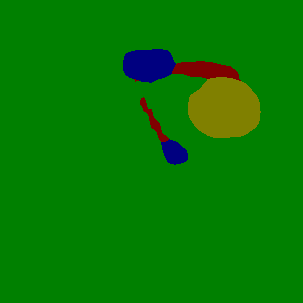}}~~\\
		\subfigure{\includegraphics[width=25mm]{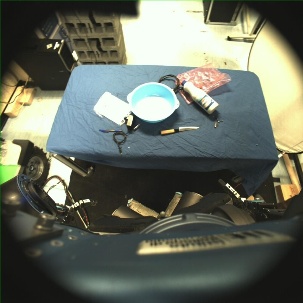}}~~
		\subfigure{\includegraphics[width=25mm]{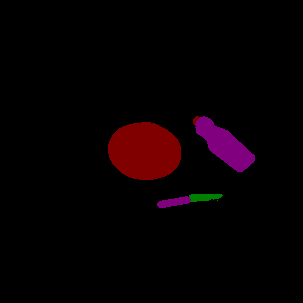}}~~
		\subfigure{\includegraphics[width=25mm]{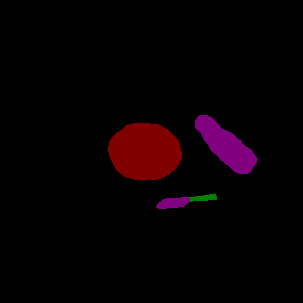}}~~
		\subfigure{\includegraphics[width=25mm]{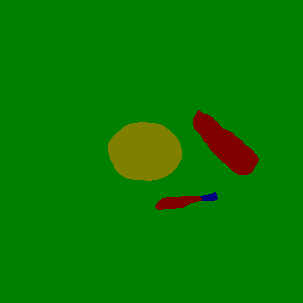}}~~\\
		\subfigure{\includegraphics[width=25mm]{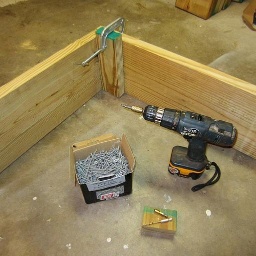}}~~
		\subfigure{\includegraphics[width=25mm]{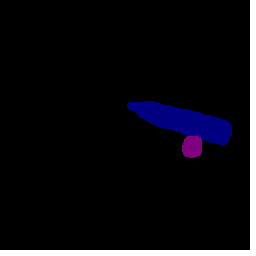}}~~
		\subfigure{\includegraphics[width=25mm]{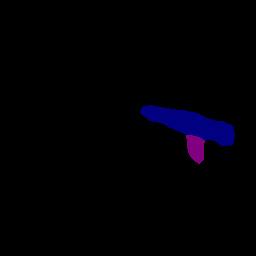}}~~
		\subfigure{\includegraphics[width=25mm]{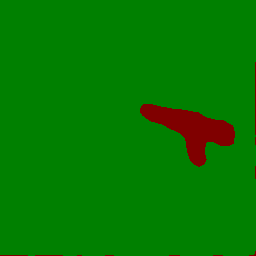}}~~\\
		\subfigure{\includegraphics[width=25mm]{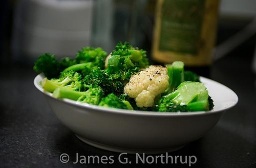}}~~
		\subfigure{\includegraphics[width=25mm]{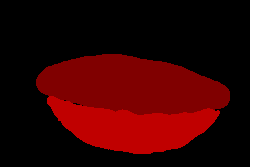}}~~
		\subfigure{\includegraphics[width=25mm]{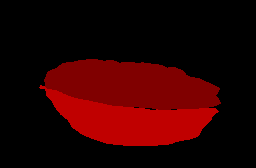}}~~
		\subfigure{\includegraphics[width=25mm]{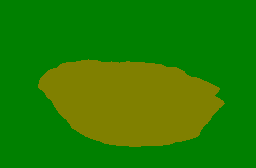}}~~\\
		\subfigure{\includegraphics[width=25mm]{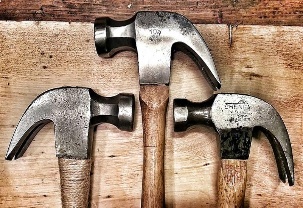}}~~
		\subfigure{\includegraphics[width=25mm]{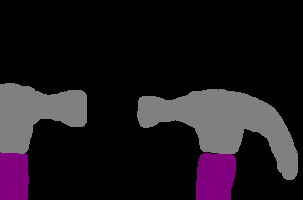}}~~
		\subfigure{\includegraphics[width=25mm]{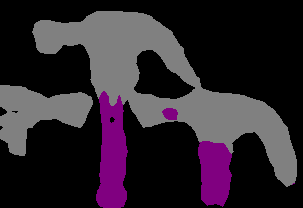}}~~
		\subfigure{\includegraphics[width=25mm]{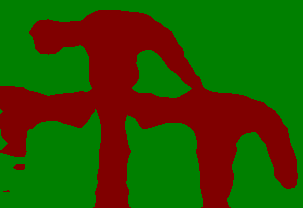}}~~\\
		\subfigure{\includegraphics[width=25mm]{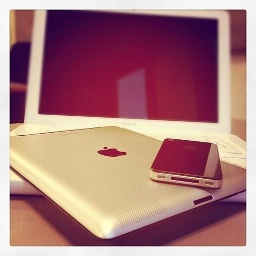}}~~
		\subfigure{\includegraphics[width=25mm]{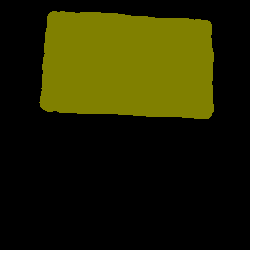}}~~
		\subfigure{\includegraphics[width=25mm]{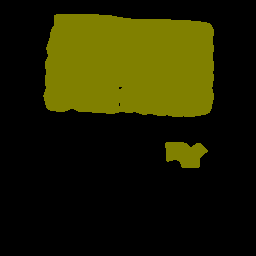}}~~
		\subfigure{\includegraphics[width=25mm]{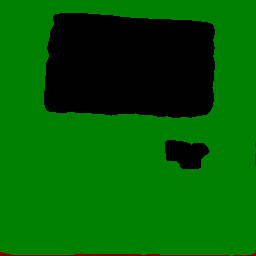}}~~\\
		\caption{Qualitative examples from the IIT Affordances test set. From left to right: image, ground truth, proposed, latent classes.}
		\label{fig:qualitative_examples_IIT}
	\end{figure*}

	Figures \ref{fig:qualitative_examples_VOC},  \ref{fig:qualitative_examplesCS} and \ref{fig:qualitative_examples_IIT} show additional qualitative results on Pascal VOC 2012, Cityscapes and IIT Affordances, respectively.
	
	\section{Manual assignment}
	Table \ref{tab:cls2cat} lists the manual assignment of the semantic classes to 10 supercategories as it is used for the experiment “manual” in Table 5 of the paper.
	\begin{table}[]
		\centering
		\caption{Manual assignment of Pascal VOC 2012 classes to 10 supercategories that we use instead of learned latent classes in the ablation study.}
		\begin{tabular}{|C{50mm}|C{50mm}|}
			\hline
			\multicolumn{2}{|c|}{Mapping of semantic classes to supercategories} \\
			\hline
			Manually defined supercategory & VOC semantic classes  \\
			\hline
			Background & Background \\
			\hline
			Aeroplane & Aeroplane \\
			\hline
			Bicycle & Bicycle \\
			\hline
			Bird & Bird \\
			\hline
			Boat & Boat \\
			\hline
			Person & Person \\
			\hline
			Ground vehicle with engine & Bus, car, motorbike, train \\
			\hline
			Mammal & Cat, cow, dog, horse, sheep \\
			\hline
			Furniture & Dinning table, sofa, chair \\
			\hline
			Miscellaneous & Bottle, tv monitor, potted plant \\
			\hline
		\end{tabular}
		\label{tab:cls2cat}
	\end{table}

\end{document}